\documentclass[twoside,11pt]{article}

\usepackage{jmlr2e, amsmath, color,array, booktabs,bm,algorithm, algorithmic}

\DeclareGraphicsExtensions{.pdf}

\usepackage{tikz}

\usetikzlibrary{arrows,shapes,backgrounds,patterns,fadings,decorations.pathreplacing,decorations.pathmorphing}
\tikzset{>=stealth'}
\tikzstyle{graphnode} = [circle,draw=black,minimum size=22pt,text centered, text width=22pt, inner sep=0pt]
\tikzstyle{var}   =[graphnode,fill=white]
\tikzstyle{obs}   =[graphnode,fill=black,text=white]
\tikzstyle{fac}   =[rectangle,draw=black,fill=black!25,minimum size=5pt]
\tikzstyle{facprior} =[rectangle,draw=black,fill=black,text=white,minimum size=5pt]
\tikzstyle{edge}  =[draw=white,double=black,thick,-]
\tikzstyle{prior} =[rectangle, draw=black, fill=black, minimum size=
5pt, inner sep=0pt]
\tikzstyle{dirprior} = [circle, draw=black, fill=black, minimum
size=5pt, inner sep=0pt]

\newcommand{\II}{\mathbb{I}}

\def\x{{\mathbf x}}
\def\y{{\mathbf y}}
\def\c{{\mathbf c}}

\def\r{{\mathbf r}}
\def\e{{\mathbf e}}
\def\l{{\mathbf l}}
\def\u{{\mathbf u}}
\def\a{{\mathbf a}}
\def\m{{\mathbf m}}
\def\boldmu{\bm{\mu}}

\def\boldtheta{\bm{\theta}}
\def\regionA{\mathcal{A}}

\newcommand{\reals}{\mathsf{I\!R}}
\newcommand{\N}{\mathcal{N}}
\renewcommand{\d}{\,\mathrm{d}}
\newcommand{\KL}{\text{KL}}

\newcommand{\wo}{\setminus}

\jmlrheading{X}{2011}{XX-XX}{04/11}{X/11}{John P. Cunningham, Philipp Hennig, and Simon Lacoste-Julien}

\ShortHeadings{Gaussian Probabilities and Expectation Propagation}{Cunningham, Hennig, Lacoste-Julien}
\firstpageno{1}

\begin{document}

\newpage

\title{Gaussian Probabilities\\ and Expectation Propagation}

\author{%
\name John P. Cunningham
\email jpc74@cam.ac.uk \\
\addr Department of Engineering\\
University of Cambridge\\
Cambridge, UK\\
\AND
\name Philipp Hennig
\email{phennig@tuebingen.mpg.de} \\
\addr Max Planck Institute for Intelligent Systems\\
T\"ubingen, Germany\\
\AND
\name Simon Lacoste-Julien
\email{sl522@cam.ac.uk} \\
\addr Department of Engineering\\
University of Cambridge\\
Cambridge, UK\\
}

\editor{Editor Name}

\maketitle

\begin{abstract}
  While Gaussian probability \emph{densities} are omnipresent in applied mathematics, Gaussian cumulative \emph{probabilities} are hard to calculate in any but the univariate case. 
  We study the utility of Expectation Propagation (EP) as an approximate integration method for this problem. For rectangular integration regions, the approximation is highly accurate. We also extend the derivations to the more general case of polyhedral integration regions. However, we find that in this polyhedral case, EP's answer, though often accurate, can be almost arbitrarily wrong. We consider these unexpected results empirically and theoretically, both for the problem of Gaussian probabilities and for EP more generally.  These results elucidate an interesting and non-obvious feature of EP not yet studied in detail.  \end{abstract}



\section{Introduction}
\label{sec:introduction}

This paper studies approximations to definite integrals of Gaussian (also known as normal or central) probability distributions. We define the Gaussian distribution $p_0(\x) = \N(\x;\m,K)$ as
\begin{equation} 
\label{eqn:gaussian} 
p_0(\x) =  \frac{1}{(2\pi)^{\frac{n}{2}}|K|^{\frac{1}{2}}} \exp\left\{ -\frac{1}{2}(\x - \m)^T K^{-1}(\x - \m)\right\}, \end{equation}
\noindent where $\x\in\reals^{n}$ is a vector with $n$ real valued elements, $\m\in\reals^{n}$ is the mean vector, and $K\in\reals^{n\times n}$ is the symmetric, positive semidefinite covariance matrix. The Gaussian is perhaps the most widely used distribution in science and engineering. Whether this popularity is reflective of a fundamental character (often afforded to this distribution because of its role in the Central Limit Theorem) is debatable, but at any rate, a battery of convenient analytic characteristics make it an indispensible tool for many applications.   Multivariate Gaussians induce Gaussian marginals and conditionals on all linear subspaces of their domain, and are closed under linear transformations of their domain. Under Gaussian beliefs, expectations can be evaluated for a number of important function classes, such as linear functions, quadratic forms, exponentials of the former, trigonometric functions, and linear combinations of such functions. Gaussians form an exponential family (which also implies closure, up to normalisation, under multiplication and exponentiation, and the existence of an analytic conjugate prior---the normal-Wishart distribution). The distribution's moment generating function and characteristic function have closed-form. Historically, much importance has been assigned to the fact that the Gaussian maximises differential entropy for given mean and covariance. More recently, the fact that the distribution can be analytically extended to the infinite-dimensional limit---the Gaussian Process \citep{rasmussenBook}---has led to a wealth of new applications. Yet, despite all these great aspects of Gaussian \emph{densities}, Gaussian \emph{probabilities} are difficult to calculate: even the cumulative distribution function (cdf) has no closed-form expression and is numerically challenging in high-dimensional spaces. More generally, we consider the probability that a draw from $p(\x)$ falls in a region $\regionA\subseteq\reals^{n}$, which we will denote as
\begin{equation}
\label{eqn:cumdensity}
F(\regionA) = \mathrm{Prob}\left\{\x\in \regionA\right\} = \int_\regionA p(\x)\d\x =
\int_{\ell_1(\x)}^{u_1(\x)} \cdots \int_{\ell_n(\x)}^{u_n(\x)} p(\x) \d
x_n \cdots \d x_1,
\end{equation}
where $\ell_1,{\ldots} ,\ell_n$ and $u_1,{\ldots} ,u_n$ denote the upper and lower bounds of the region $\regionA$.  In general, $\u(\x)$ and $\l(\x)$ may be nonlinear functions of $\x$ defining a valid region $\regionA$, but the methods presented here discuss the case where $\u$ and $\l$ are linear functions of $\x$, meaning that $\regionA$ forms a (possibly unbounded) polyhedron. The probability $F(\regionA)$ generalises the cdf, which is recovered by setting the upper limits $\u=(u_i)_{i=1,\dots,n}$ to a single point in $\reals^n$ and the lower limits to $l_1={\ldots} =l_n = -\infty$. Applications of these multivariate Gaussian probabilities are widespread. They include statistics \citep{genz92, joeMVNcdfJASA1995, Hothorn05unbiasedrecursive} (where a particularly important use case is probit analysis \citep{AshfordSowden1970,SicklesTabuman1986,Gibbons1996}), economics \citep{Boyle05pricingoptions}, mathematics \citep{Hickernell99theasymptotic}, biostatistics \citep{thiebautCMPB2004,zhaoCSDA2005}, medicine \citep{lesaffre1991}, environmental science \citep{Buccianti_computationalinvestigations}, computer science \citep{Klerk_onapproximate}, neuroscience \citep{Pillow04maximumlikelihood}, machine learning \citep{liaoICML2007}, and more (see for example the applications listed in \citet{gassmann2002}).  For the machine learning researcher, two popular problems that can be cast as Gaussian probabilities include the Bayes Point Machine \cite[]{herbrichBook} and Gaussian Process classification \cite[]{rasmussenBook, KussRasmussen2005}, which will we discuss more specifically later in this work.

Univariate Gaussian probabilities can be so quickly and accurately calculated \citep[\emph{e.g.}][]{cody69} that the univariate cumulative density function is available with machine-level precision in many statistical computing packages ({\it e.g.}, {\tt normcdf} in {\sc matlab}, {\tt CDFNORM} in {\sc spss}, {\tt pnorm} in {\sc r}, to name a few).  Unfortunately, no similarly powerful algorithm exists for the multivariate case. There are some known analytic decompositions of Gaussian integrals into lower-dimensional forms \citep{Placket1954,Curnow1962,Lazard2003}, but such decompositions have very high computational complexity \citep{Huguenin2009}. In statistics, sampling based methods have found application in specific use cases \citep{LermanManski1981,McFadden1989,Pakes1989}, but the most efficient general, known method is numerical integration \citep{genz92, drezner89, drezner94, genz99, genz99b, genz02, genz04}. A recent book \citep{genzBook} gives a good overview. These algorithms make a series of transformations to the Gaussian $p_0(\x)$ and the region $\regionA$, using the Cholesky factor of the covariance $K$, the univariate Gaussian cdf and its inverse, and randomly generated points.  These methods aim to restate the calculation of $F(\regionA)$ as a problem that can be well handled by quasi-random or lattice point numerical integration. Many important studies across many fields have been critically enabled by these and other algorithms \citep{joeMVNcdfJASA1995, Hothorn05unbiasedrecursive, Boyle05pricingoptions, Hickernell99theasymptotic, thiebautCMPB2004, zhaoCSDA2005, Buccianti_computationalinvestigations, Klerk_onapproximate, Pillow04maximumlikelihood, liaoICML2007}.   In particular, the methods of Genz represent the state of the art, and we will use this method as a basis for comparison.  We note that the numerical accuracy of these methods is generally high and can be increased by investing additional computational time (more integration points), though by no means to the numerical precision of the univariate case.  As such, achieving high accuracy invokes substantial computational cost.  Also, applications such as Bayesian model selection require \emph{analytic} approximations of $F(\regionA)$, usually because the goal is to optimise $F(\regionA)$ with respect to the parameters $\{\m,K\}$ of the Gaussian, which requires the corresponding derivatives.  These derivatives and other features (to be discussed) are not currently offered by numerical integration methods.  Thus, while there exist sophisticated methods to compute Gaussian probabilities with high accuracy, there remains significant work to be done to address this important problem.  

In Section \ref{sec:ep}, we will develop an analytic approximation to $F(\regionA)$ by using \emph{Expectation Propagation (EP)} \cite[]{minka01phd, minkaUAI01,minkaMSFTTR2005, opperTAP2000} as an approximate integration method. We first give a brief introduction to EP (Section \ref{sec:expect-prop}), then develop the necessary methods for integration over \emph{hyperrectangular} regions $\regionA$ in Section \ref{sec:rect-integr-regi}, which is the case most frequently computed (and which includes the cdf).  In Section \ref{sec:polyh-integr-regi}, we then describe how to generalise these derivations to \emph{polyhedral} $\regionA$. However, while these polyhedral extensions are conceptually elegant, it turns out that they do not always lead to an accurate algorithm. In Section \ref{sec:results}, we compare EP's analytic approximations to numerical results. For rectangular $\regionA$, the approximations are of generally high quality. For polyhedral $\regionA$, they are often considerably worse. This shortcoming of EP will come as a surprise to many readers, because the differences to the rectangular case are inconspiciously straightforward.  Indeed, we will demonstrate that hyperrectangles are in fact not a fundamental distinction, and we will use intuition gained from the polyhedral case to build pathological cases for hyperrectangular cases also.  We study this interesting result in Section \ref{sec:discussion} and give some insight into this problem. Our overall empirical result is that EP provides reliable analytic approximations for rectangular Gaussian integrals, but should be used only with caution on more general regions. This has implications for some important applications, such as Bayesian generalised regression, where EP is often used.

This work bears connection to previous literature in machine learning.  Most obviously, when considering the special case of Gaussian probabilities over hyperrectangular regions (Section \ref{sec:rect-integr-regi}), our algorithm is derived directly from EP, yet distinct in two ways.  First, conceptually, we do not use EP for approximate inference, but instead we put integration bounds in the place of likelihood terms, thereby using EP as a high dimensional integration scheme.  Second, because of this conceptual change, we are dealing with unnormalised EP (as the ``likelihood" factors are not distributions).  From this special case, we extend the method to more general integration regions, which involves rank one EP updates.  Rank one EP updates have been previously discussed, particularly in connection to the Bayes Point Machine \cite[]{minkaUAI01, herbrichBook, minkaTR2008}, but again their use for Gaussian probabilities has not been investigated. 

Thus, our goal here is to build from EP and the general importance of the Gaussian probability problem, to offer three contributions: first, we give a full derivation that focuses EP on the approximate integration problem for multivariate Gaussian probabilities.  Second, we perform detailed numerical experiments to benchmark our method and other methods, so that this work may serve as a useful reference for future researchers confronted with this ubiquitous and challenging computation.  Third, we discuss empirically and theoretically some features of EP that have not been investigated in the literature, which has importance for EP well beyond Gaussian probabilities.

The remainder of this paper is laid out as follows.  In Section \ref{sec:ep}, we discuss the EP algorithm in general and its specific application when the approximating distribution is Gaussian, as is our case of interest here.  In Section \ref{sec:epmgp}, we apply EP specifically to Gaussian probabilities, describing the specifics and features of our algorithm: Expectation Propagation for Multivariate Gaussian Probabilities (EPMGP).  In Section \ref{sec:othermgp}, we describe other existing methods for calculating Gaussian probabilities, as they will be compared to EPMGP in the results section.  In Section \ref{sec:testcases}, we describe the probabilities that form the basis of our numerical computations.  Section \ref{sec:results} gives the results of these experiments, and Section \ref{sec:discussion} discusses the implications of this work and directions for future work.

\section{Approximate Integration Through Approximate Moment Matching: Expectation Propagation}
\label{sec:ep}

The key step to motivate this approach is to note that we can cast the Gaussian probability problem as one of integrating an intractable and unnormalised distribution.  Defining $p(\x)$ as 
\begin{gather}
\label{eqn:trunc2}
p(\x) = \begin{cases} 
p_0(\x) & \x \in \regionA \\
~~0~~ & \mathrm{otherwise,} 
\end{cases}
\end{gather}

\noindent we see that the normaliser of this distribution is our probability of interest $F(\regionA) = \mathrm{Prob}\left\{\x\in \regionA\right\} = \int_\regionA p_0(\x)\d\x = \int p(\x) \d\x$.

Expectation Propagation \cite[]{minka01phd, minkaUAI01,minkaMSFTTR2005, opperTAP2000} is a method for finding an approximate unnormalised distribution $q(\x)$ to replace an intractable true unnormalised distribution $p(\x)$. The EP objective function is involved, but EP is \emph{motivated} by the idea of minimising Kullback-Leibler (KL) divergence \cite[]{CoverandThomas} from the true distribution to the approximation. We define this KL-divergence as 
\begin{equation}
  \label{eq:1}
  D_\KL(p \| q) = \int p(\x) \log \frac{p(\x)}{q(\x)} \d \x + \int q(\x) \d \x -
  \int p(\x) \d \x
\end{equation}
which is a form of the divergence that allows unnormalised distributions and reverts to the more popular form if $p$ and $q$ have the same indefinite integral. 

KL-divergence is a natural measure for the
quality of the approximation $q(\x)$.  If we choose $q(\x)$ to be
a high-dimensional Gaussian, then the choice of $q(\x)$ that
minimises $D_\KL(p \parallel q)$ is
the $q(\x)$ with the same zeroth, first, and
second moments as $p(\x)$.  If we are trying
to calculate $F(\regionA)$, then we
equivalently seek the zeroth moment of $p(\x)$.  As such, trying to minimise this global KL
divergence is an appropriate and sensible method to calculate Gaussian probabilities.  Appendix~\ref{sec:KL} reviews the equivalence between moment matching and minimising KL-divergence in this problem. 

 Unfortunately, minimising
global KL-divergence directly is in many cases intractable.  This
fact motivated the creation of the EP algorithm, which seeks to approximately do
this global KL minimisation by
iteratively minimising the KL-divergence of local, single factors
of $q(\x)$ with respect to $p(\x)$.  Since $F(\regionA)$ is the zeroth moment of $p(\x)$ as defined in Equation (\ref{eqn:cumdensity}), any algorithm trying to minimise this global KL-divergence objective, or an approximation thereof, is also a candidate method for the (approximate) calculation of Gaussian probabilities.

\subsection{Expectation Propagation}
\label{sec:expect-prop}

The following sections review EP and introduce its use for Gaussian probability calculations.  It is a prototypical aspect of Bayesian inference that an intractable distribution $p(\x)$ is a product of a prior distribution $p_0(\x)$ and one or more likelihood functions or factors $t_i(\x)$: 

\begin{equation}
  \label{eq:2}
  p(\x) = p_0(\x) \prod_i t_i(\x).
\end{equation}

\noindent Note that our Gaussian probability problem has this unnormalised form, as it can be written as: 

\begin{equation}
  \label{eqn:cumdensityEP}
  F(\regionA) = \int_\regionA p_0(\x)\d\x = \int p(\x)\d\x = \int p_0(\x) \prod_i t_i(\x) \d\x
\end{equation}
where $t_i(\x)$ is an indicator function defined in a particular direction, namely a ``box function":

\begin{equation}
\label{eqn:truefactor}
t_i(\x)  =  \II \bigl\{ l_i < \c_i^T\x < u_i \bigr\} = 
\begin{cases}
1 & ~~ l_i < \c_i^T\x < u_i \\
0 & ~~ \mathrm{otherwise}. 
\end{cases}
 \end{equation}

The above form is our intractable problem of interest\footnote{This box function notation may seem awkward vs. the more conventional halfspace definitions of polyhedra, but we will make use of this definition.  Further, halfspaces can be recovered by setting any $u_i = \infty$, so this definition is general.}, and we assume without loss of generality that the $\c_i$ have unit norm.  An important clarification for the remainder of this work is to note that most of the distributions discussed will be unnormalised.  By the previous construction, the factors $t_i(\x)$ are unnormalised, and thus so is $p(\x)$ (as will be the EP approximation $q(\x)$).  Though the steps we present will all be valid for unnormalised distributions, we make this note to call out this somewhat atypical problem setting.

   The approach of the EP algorithm is to replace each intractable $t_i(\x)$ with a tractable unnormalised Gaussian $\tilde{t}_i(\x)$.  Our specific algorithm thus yields a nice geometric interpretation.  We want to integrate a Gaussian over a polyhedron defined by several box functions, but this operation is intractable.  Instead, EP allows us to replace each of those intractable box truncation functions with soft Gaussian truncations $\tilde{t}_i(\x)$.   Then, since we know that these exponential family distributions are simple to multiply (multiplication essentially amounts to summing the natural parameters), our problem reduces to finding $\tilde{t}_i(\x)$.

This perspective motivates the EP algorithm, which approximately
minimises $D_\KL(p \| q)$ by iteratively constructing an unnormalised approximation
$q(\x)$. At any point in the
iteration, EP tracks a current approximate $q(\x)$ in an
exponential family, and a set of \emph{approximate factors} (or \emph{messages}) $\tilde{t}_i$, also
in the family. The factors are updated by constructing a
\emph{cavity} distribution
\begin{equation}
  \label{eq:5}
  q^{\wo i}(\x) = \frac{q(\x)}{\tilde{t}_i(\x)}
\end{equation}
(this division operation on unnormalised distributions is also well-defined for members
of exponential families - it amounts to subtracting natural parameters), and then
\emph{projecting} into the exponential family
\begin{equation}
  \label{eq:6}
  \tilde{t}_i(\x)q^{\wo i}(\x) = \operatorname{proj} [t_i(\x)q^{\wo i}(\x)].
\end{equation}
where this projection operation (\cite{PowerEP}, an M-projection from information geometry \cite[]{kollerBook}) is defined as setting $\tilde{t}_i(\x)$ to the unnormalised member of the exponential family minimizing $D_\KL(t_i q^{\wo i}\| \tilde{t}_i q^{\wo i})$.   Intuitively, to update an approximate factor $\tilde{t}_i(\x)$, we first remove its effect from the current approximation (forming the cavity), and then we include the effect of the true factor $t_i(\x)$ by an M-projection, updating $\tilde{t}_i(\x)$ accordingly.  As derived in Appendix~\ref{sec:KL}, this projection means matching the sufficient statistics of $\tilde{t}_i(\x)q^{\wo i}(\x)$ to those of $t_i(\x) q^{\wo i}(\x)$. In particular for Gaussian $\tilde{t}_i(\x)$, matching sufficient statistics is equivalent to matching \emph{zeroth}, first and second moments.   We now restrict this general prescription of EP to the Gaussian case, as it will clarify the simplicity of our Gaussian probability algorithm.

\subsubsection{Gaussian EP with rank-one factors}
\label{sec:gaussian-ep}

In cases where the approximating family is Gaussian, the EP derivations can be specified further without having to make additional assumptions about the true $t_i(\x)$. Because it will lend clarity to Sections \ref{sec:rect-integr-regi} and \ref{sec:polyh-integr-regi}, we give that analysis here.  To orient the reader, it is important to be explicit about the numerous uses of ``Gaussian" here.  When we say ``Gaussian EP," we mean \emph{any} EP algorithm where the approximating $q(\x)$ (the prior $p_0(\x)$ and approximate factors $\tilde{t}_i(\x)$) are Gaussian.  The true factors $t_i(\x)$  can be arbitrary.  In particular, here we discuss Gaussian EP, and only in Section \ref{sec:epmgp} do we discuss Gaussian EP as used for multivariate Gaussian probability calculations (with box function factors $t_i(\x)$).  It is critical to note that all steps in this next section are invariant to changes in the form of the true $t_i(\x)$, so what follows is a general description of Gaussian EP.  The unnormalised Gaussian approximation is 
 
\begin{equation}
q(\x) = p_0(\x)\prod_i \tilde{t}_i(\x) = p_0(\x)\prod_i \tilde{Z}_i\mathcal{N}(\x; \tilde{\mu}_i,\tilde{\sigma}_i^2) = Z \N(\x; \mu, \Sigma).
\end{equation}

We will require the cavity
distribution for the derivation of the updates, which is defined as

\begin{equation}
q^{\wo i}(\x) = \frac{q(\x)}{\tilde{t}_i(\x)} = Z^{\wo i} \N(\x ; \u^{\wo i} , V^{\wo i})
\end{equation}

The above step is the \emph{cavity} step of EP (Equation \ref{eq:5}).  We now must do the projection operation of Equation \ref{eq:6}, which involves moment matching the approximation $\tilde{t}_i(\x)q^{\wo i}(\x)$ to the appropriate moments of $t_i(\x)q^{\wo i}(\x)$:

\begin{eqnarray}
\label{eqn:moments}
\hat{Z}_i & = & \int t_i(\x) q^{\wo i}(\x) \d\x  \\
\hat{\u}_i & = & \frac{1}{\hat{Z}_i}\int \x t_i(\x) q^{\wo i}(\x) \d\x  \\
\hat{V}_i & = & \frac{1}{\hat{Z}_i}\int (\x - \hat{\u}_i)(\x - \hat{\u}_i)^T t_i(\x) q^{\wo i}(\x) \d\x. 
\end{eqnarray}

These updates are useful and tractable if the individual $t_i(\x)$ have some simplifying structure such as low rank, as in Equation \ref{eqn:truefactor}.  In particular, as is the case of interest here, if the $t_i(\x)$ have rank one structure, then these moments have simple form:

\begin{eqnarray}
\label{eqn:r1moments}
\hat{Z}_i & = & \int t_i(\x) q^{\wo i}(\x) \d\x  \nonumber \\
\hat{\u}_i & = & \hat{\mu}_i   \c_i \nonumber \\
\hat{V}_i & = & \hat{\sigma}_i^2 \c_i \c_i^T,
\end{eqnarray}

\noindent where $\{\hat{Z}_i,\hat{\mu}_i,\hat{\sigma}_i^2\}$ depend on the factor $t_i(\x)$ and $\c_i$ is the rank one direction as in Equation \ref{eqn:truefactor}.   The important thing to note is that these parameters $\{\hat{Z}_i,\hat{\mu}_i,\hat{\sigma}_i^2\}$ are the only place in which the actual form of the $t_i(\x)$ enter in.   This can also be derived from differential operators, as was done in a short technical report \cite[]{minkaTR2008}.

To complete the projection step, we must update the approximate factors $\tilde{t}_i(\x)$ such that the moments of $\tilde{t}_i(\x)q^{\wo i}(\x)$ match $\{\hat{Z}_i,\hat{\mu}_i,\hat{\sigma}_i^2\}$, as doing so is the KL minimisation over this particular factor $\tilde{t}_i(\x)$.  We derive the simple approximate factor updates using basic properties of the Gaussian (again just subtracting natural parameters).  We have

\begin{eqnarray}
\label{eqn:site}
 \tilde{t}_i(\x) & = &
\tilde{Z}_i\mathcal{N}(\c_i^T\x; \tilde{\mu}_{i},\tilde{\sigma}^2_{i}),~~~\\
\mathrm{where}~~~\tilde{\mu}_{i} & = &
\tilde{\sigma}^2_{i}(\hat{\sigma}^{-2}_i\hat{\mu}_i -
{\sigma}^{-2}_{\wo i}{\mu}_{\wo i}),~~~~~~
\tilde{\sigma}^2_{i} = (\hat{\sigma}^{-2}_i -
\sigma^{-2}_{\wo i})^{-1},\nonumber\\
\tilde{Z}_i & = & \hat{Z}_i\sqrt{2\pi}\sqrt{\sigma^2_{\wo i} +
\tilde{\sigma}^2_i}\mathrm{exp}\Bigl\{\frac{1}{2}(\mu_{\wo i} -
\tilde{\mu}_i)^2/(\sigma^2_{\wo i} + \tilde{\sigma}_i^2)\Bigr\},\nonumber
\end{eqnarray}

\noindent and where cavity parameters $\{\sigma_{\wo i}^2,\mu_{\wo i} \} ~= ~\{ \c_i^TV^{\backslash i}\c_i ~,~ \c_i^T\u^{\backslash i} \}$ are fully derived in Appendix \ref{sec:rank1cavity} as:

\begin{eqnarray}
\label{eqn:cavityepmgp}
\mu_{\wo i} = \sigma_{\wo i}^2 \Bigl( \frac{\c_i^T\boldmu}{\c_i^T\Sigma\c_i} - \frac{\tilde{\mu}_i}{\tilde{\sigma}_i^2} \Bigr)
,~~~\mathrm{and}~~~
\sigma^2_{\wo i}  = \bigl( (\c_i^T\Sigma \c_i)^{-1} - \tilde{\sigma}^{-2}_i\bigr)^{-1}.
\end{eqnarray}

\noindent These
equations are all standard manipulations of Gaussian random variables.   Now, by the definition of the approximation, we can calculate the new approximation $q(\x)$ as the product $q(\x) = p_0(\x)\prod_i \tilde{t}_i(\x)$:

\begin{eqnarray}
\label{eqn:epmgppost}
 q(\x) & = &
Z\mathcal{N}(\boldmu,\Sigma),~~~\\
\mathrm{where}~~~\boldmu & = &
\Sigma\Bigl(K^{-1}\m + \overset{m}{\sum_{i=1}}\frac{\tilde{\mu}_i}{\tilde{\sigma}^2_i}\c_i \Bigr),~~~~~~\Sigma = \Bigl(K^{-1} +
\overset{m}{\sum_{i=1}} \frac{1}{\tilde{\sigma}^2_i}\c_i\c_i^T\Bigr)^{-1},~~~~~~\nonumber
\end{eqnarray}

\noindent where $p_0(\x) = \mathcal{N}(\x; \m,K)$ as in Equation \ref{eqn:gaussian}.  These forms can be efficiently and numerically stably calculated in quadratic computation time via steps detailed in Appendix \ref{sec:epmgpstable}.  Lastly, once the algorithm has converged, we can calculate the normalisation constant of $q(\x)$.  While this step is again general to Gaussian EP, we highlight it because it is our quantity of interest - the approximation of the probability $F(\regionA)$:

\begin{eqnarray}
\label{eqn:logZ}
\log Z & = &
-\frac{1}{2}\left(\m^TK^{-1}\m
+\log\lvert K \rvert \right) \nonumber \\
& &
+ \overset{m}{\sum_{i=1}}\left( \log \tilde{Z}_i
- \frac{1}{2}\left(\frac{\tilde{\mu}_i^2}{\tilde{\sigma}_i^2}
+ \log \tilde{\sigma}^2_i + \log(2\pi)\right)\right) \nonumber \\
& &
+ \frac{1}{2}\left(\boldmu^T\Sigma^{-1}\boldmu
+ \log\lvert \Sigma \rvert \right)
\end{eqnarray}

\noindent In the above we have broken this equation up into three lines to clarify that the normalisation term $\log Z$ has contribution from the prior $p_0(\x)$ (first line), the approximate factors (the second line), and the full approximation $q(\x)$ (third line).  An intuitive interpretation is that the integral of $q(\x)$ is equal to the product of all the contributing factor masses (the prior and the approximate factors $\tilde{t}_i$), divided by what is already counted by the normalisation constant of the final approximation $q(\x)$.  Similar reparameterisations of these calculations help to ensure numerical stability and quicker, more accurate computation.  We derive that reparameterisation in the second part of Appendix \ref{sec:epmgpstable}.   Taking stock, we have derived Gaussian EP in a way that requires only rank one forms.  These factors may be of any functional form, but in the following section we show the box function form of the Gaussian probability problem leads to a particularly fast and accurate computation of Gaussian EP.

At this point it may be helpful to clarify again that EP has two levels of approximation.  First, being an approximate inference framework, EP would ideally choose an approximating $q(\x)$ from a tractable family (such as the Gaussian) such that $D_\KL(p\| q)$ is minimised.  That global problem being intractable, EP makes a second level of approximation in not actually minimising global $D_\KL(p\| q)$, but rather minimising the local $D_\KL(t_i q^{\wo i}\| \tilde{t}_i q^{\wo i})$. Because there is no obvious connection between these global and local divergences, it is difficult to make quantitative statements about the quality of the approximation provided by EP without empirical evaluation. However, EP has been shown to provide excellent approximations in many applications \citep{minka01phd, KussRasmussen2005, herbrich2007trueskilltm, stern2009matchbox}, often outperforming competing methods. These successes have deservedly established EP as a useful tool for applied inference, but they may also mislead the applied researcher into trusting its results more than is warranted by theoretical underpinnings. This paper, aside from developing a tool for approximate Gaussian integration, also serves to demonstrate a non-obvious way in which EP can yield poor results. The following two sections introduce our use of EP for approximate Gaussian integration that performs very well (Sections \ref{sec:rect-integr-regi} and \ref{sec:results}), followed by an ostensibly straightforward extension  (Section \ref{sec:polyh-integr-regi}), which turns out not to perform particularly well (Section \ref{sec:results}), for a subtle reason (Section \ref{sec:discussion}).

\section{EP Multivariate Gaussian Probability (EPMGP) Algorithm}
\label{sec:epmgp}

This section introduces the EPMGP algorithm, which uses Expectation Propagation to calculate multivariate Gaussian probabilities.  We consider two cases: axis-aligned hyperrectangular integration regions $\regionA$ and more general poyhedral integration regions $\regionA$.   Because of our general Gaussian EP derivation above, for each case we need only calculate the parameters that are specific to the choices of factor $t_i(\x)$, namely  $\{\hat{Z}_i,\hat{\mu}_i,\hat{\sigma}_i^2\}$ from Equation \ref{eqn:r1moments}.  This approach simplifies the presentation of this section and clarifies what is specific to Gaussian EP and what is specific to the Gaussian probability problem.

\subsection{Rectangular Integration Regions}
\label{sec:rect-integr-regi}

\begin{figure}[ht]
  \centering
  \begin{tikzpicture}[node distance=2cm]
    \node[fac] (p) at (0,0) {}; 
    \node[anchor=east] at (p.west) {$p_0(\x) = \mathcal{N}(\x; \m, K) $}; 
    \node[var, right of=p] (x) {$\x$} edge (p);
    \node[fac, right of=x] (t) {} edge (x);
    \node[anchor=west] at (t.east) {$t_i(\x)= \II\{l_i < x_i < u_i\}$};
    \draw (3,-1) rectangle (9,1);
    \node[anchor=south east] at (9,-1) {$i=1,\dots,n$};
  \end{tikzpicture}
  \caption{Graphical model (factor graph) for the rectangular Gaussian integration
    problem. In the language of Bayesian inference, the unconstrained
    Gaussian forms a ``prior'', the integration limits form
    ``likelihood'' factors, which give a ``posterior'' distribution on
    $\x$, and we are interested in the normalisation constant of that
    distribution. Note, however, that this inference terminology is
    used here without the usual philosophical interpretations of prior
    knowledge and subsequently aquired data.  Note also that in the rectangular integration case there are exactly $n$ such factors, corresponding to each dimension of the problem.}
  \label{fig:fac_rect}
\end{figure}
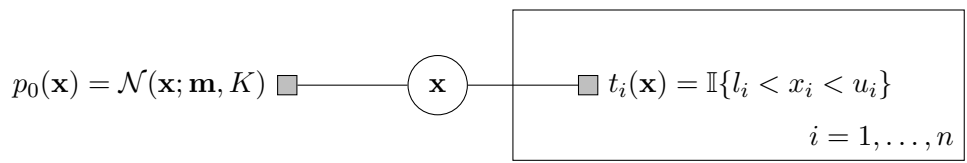

This section derives an EP algorithm for Gaussian integration over rectangular regions. Without loss of generality, we will assume that the rectangular region $\regionA$ is axis-aligned. Otherwise, there exists a unitary map $ U$ such that in the rotated coordinate system $\x\to U\x$ the region is axis aligned and the transformation morphs the Gaussian distribution $p_0(\x)=\N(\x; \m,K)$ into another Gaussian $p_0'(\x) = \N(U\x; U\m,UKU^T)$. Hence, the integration boundaries are not functions of $\x$, and the integral $F(\regionA)$ can be written using a product of $n$ factors $t_i(x_i) = \II\{l_i<x_i<u_i\}$: 

\begin{equation}
  \label{eq:7}
  F(\regionA) = \int_{l_1} ^{u_1} \cdots \int_{l_n} ^{u_n} \N(\x;\m,K) \d\x = \int \N(\x;\m,K) \prod_{i=1} ^n t_i(x_i) \d x_i.
\end{equation}

EP's iterative steps thus correspond to the incorporation of individual box functions of upper and lower integration limits into the approximation, where the incorporation of a factor implies approximating the box function with a univariate Gaussian aligned in that direction.  Thanks to the generality of the derivations in Section \ref{sec:gaussian-ep}, the remaining derivations for this setting are conveniently short.  The required moments for Equation \ref{eqn:r1moments} (the EP projection step) are calculated exactly \cite[]{jawitz2004} using the error function\footnote{These erf($\cdot$) computations also require a numerically stable implementation, though the details do not add insight to this problem and are not specific to this problem (unlike Appendix \ref{sec:epmgpstable}).  Thus we defer those details to the EPMGP code package available at {\tt <url to come with publication>}.}:
\begin{eqnarray}
\label{eqn:hat}
\hat{Z}_i & = &
\frac{1}{2}\Bigl(\mathrm{erf}(\beta) -  \mathrm{erf}(\alpha)\Bigr), \\
\hat{\mu}_i & = & \mu_{\wo i} +
\frac{1}{\hat{Z}_i}\frac{\sigma_{\wo i}}{\sqrt{2\pi}}\bigl(\mathrm{exp}\{-\alpha^2\}
- \mathrm{exp}\{-\beta^2\}\bigr),~~~\mathrm{and} \\
\label{eqn:hat3}
\hat{\sigma}_i^2 & = &   \mu_{\wo i}^2 + \sigma_{\wo i}^2 +
\frac{1}{\hat{Z}_i}\frac{\sigma_{\wo i}}{\sqrt{2\pi}}\bigl((l_i +
\mu_{\wo i})\mathrm{exp}\{-\alpha^2\} - (u_i +
\mu_{\wo i})\mathrm{exp}\{-\beta^2\}\bigr) - \hat{\mu}_i^2, 
\end{eqnarray}
\noindent where we have everywhere above used the shorthand $\alpha = \frac{l_i -
\mu_{\wo i}}{\sqrt{2}\sigma_{\wo i}}$ and $\beta = \frac{u_i -
\mu_{\wo i}}{\sqrt{2}\sigma_{\wo i}}$, and  $\{\mu_{\wo i},\sigma^2_{\wo i}\}$ are:

\begin{eqnarray}
\label{eqn:cavityepmgpAxis}
\mu_{\wo i} & = &  \sigma_{\wo i}^2 \Bigl( \frac{\mu_i}{\Sigma_{ii}} - \frac{\tilde{\mu}_i}{\tilde{\sigma}_i^2} \Bigr)
,~~~\mathrm{and}~~~
\sigma^2_{\wo i} = \bigl( \Sigma_{ii}^{-1} - \tilde{\sigma}^{-2}_i\bigr)^{-1}.
\end{eqnarray}

The above is simply the form of  Equation \ref{eqn:cavityepmgp} with the vectors $\c_i$ set to the cardinal axes $\e_i$.  These parameters are the forms required for Gaussian EP to be used for hyperrectangular Gaussian probabilities.

\subsection{Polyhedral Integration Regions}
\label{sec:polyh-integr-regi}

\begin{figure}[ht]
  \centering
  \begin{tikzpicture}[node distance=2cm]
    \node[fac] (p) at (0,0) {}; 
    \node[anchor=east] at (p.west) {$p_0(\x) = \mathcal{N}(\x; \m, K) $}; 
    \node[var, right of=p] (x) {$\x$} edge (p);
    \node[fac, right of=x] (t) {} edge (x);
    \node[anchor=west] at (t.east) {$t_i(\x)= \II\{l_i < \c_i^T\x < u_i\}$};
    \draw (3,-1) rectangle (9,1);
    \node[anchor=south east] at (9,-1) {$i=1,\dots,m$};
  \end{tikzpicture}
  \caption{Factor graph for the polyhedral Gaussian integration
    problem. Note the onstensibly very similar setup to the rectangular problem in Figure \ref{fig:fac_rect}. The only difference is a replacement of axis-aligned box functions with box functions on the axis defined by $\c_i$, and a change to the number of restrictions, from $n$ to some different $m$.}
  \label{fig:fac_polyhedral}
\end{figure}
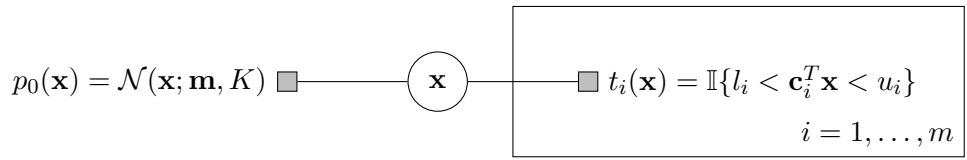

The previous section assumed axis-aligned integration limits. While a fully general set of integration limits is difficult to analyse, the derivations of the previous section can be extended to \emph{polyhedral} integration regions. This extension is straightforward: each factor truncates the distribution along the axis orthogonal to the vector $\c_i$ (which we assume has unit norm without loss of generality) instead of the cardinal axes as in the previous section.  The notational overlap of $\c_i$ with the Gaussian EP derivation of Section \ref{sec:gaussian-ep} is intentional, as these factors of Figure \ref{fig:fac_polyhedral} are precisely our general rank one factors for EP. An important difference to note in Figure \ref{fig:fac_polyhedral} is that there are $m$ factors, where $m$ can be larger or smaller (or equal to) the problem dimensionality $n$, unlike in the hyperrectangular case where $m=n$.  To implement EP, we find the identical moments as shown in Equation \ref{eqn:hat}. In fact the only difference is the cavity parameters $\{\mu_{\wo i},\sigma^2_{\wo i}\}$, which have the general form of Equation \ref{eqn:cavityepmgp}:

  \begin{eqnarray}
\label{eqn:cavityepmgpPoly}
\mu_{\wo i} & = &  \sigma_{\wo i}^2 \Bigl( \frac{\c_i^T\boldmu}{\c_i^T\Sigma\c_i} - \frac{\tilde{\mu}_i}{\tilde{\sigma}_i^2} \Bigr)
,~~~\mathrm{and}~~~
\sigma^2_{\wo i} = \bigl( (\c_i^T\Sigma \c_i)^{-1} - \tilde{\sigma}^{-2}_i\bigr)^{-1}.
\end{eqnarray}

From this point of view, it might seem like there is very little difference between the axis-aligned hyperrectangular setup of the previous section and the more general polyhedral setup of this section.  However, our experiments in Section \ref{sec:results} indicate that EP can have very different performance between these two cases, a finding discussed at length in subsequent sections.

\subsection{Implementation}

To summarise, we have an EP algorithm that calculates $q(\x) = Z\mathcal{N}(\x; \boldmu, \Sigma)$ such that $D_{KL}(p \| q)$ is hopefully small.  Though there is no proof for when EP methods converge to the global KL minimiser, our results (in the following sections) suggest that EPMGP does converge to a fixed point reasonably close to this optimum in the hyperrectangular integration case.  In the general polyhedral case, performance is often excellent but can also be considerably worse, such that results in this case should be treated with caution.   The EPMGP algorithm typically converges in fewer than 10 EP iterations (of all $m$ factors) and is insensitive to initialisation and factor ordering.   EPMGP pseudocode is given in Algorithm \ref{alg:epmgp}, and our MATLAB implementation is available at {\tt <url to come with publication>}.
%
\begin{algorithm}
\caption{EPMGP: Calculate $F(\regionA)$, the probability of $p_0(\x)$
on the region $\regionA$.}
\label{alg:epmgp}
\begin{algorithmic}[1]
\STATE Initialise with any $q(\x)$ defined by $Z,\boldmu,\Sigma$ (typically the parameters of $p_0(\x)$).
\STATE Initialise messages $\tilde{t}_{i}$ with zero precision.
\WHILE{$q(\x)$ has not converged}
\FOR{$i\gets 1:m$}
\STATE form cavity local $q_{\wo i}(\x)$ by Equation~\ref{eqn:cavityepmgp} (stably calculated by \ref{eqn:cavityepmgpnatural}).
\STATE calculate moments of $q_{\wo i}(\x)t_i(\x)$ by Equations~\ref{eqn:hat}-\ref{eqn:hat3}.
\STATE choose $\tilde{t}_i(\x)$ so $q_{\wo i}(\x)\tilde{t}_i(\x)$
matches above moments by Equation~\ref{eqn:site}.
\ENDFOR
\STATE update $\boldmu,\Sigma$ with new $\tilde{t}_i(\x)$ (stably using Equations~\ref{eqn:sigmanew} and \ref{eqn:munew}).
\ENDWHILE
\STATE calculate $Z$, the total mass of $q(\x)$ using Equation \ref{eqn:logZ} (stably using Equation~\ref{eqn:logZstable}).
\STATE \textbf{return} $Z$, the approximation of $F(\regionA)$.
\end{algorithmic}
\end{algorithm}

\subsection{Features of the EPMGP algorithm}
\label{sec:advantages}

Before evaluating the numerical results of EPMGP, it is worth noting a few attractive features of the EP approach, which may motivate its use in many applied settings.

\subsubsection{Derivatives with respect to the distribution's parameters}
\label{sec:deriv-with-resp}

A common desire in machine learning and statistics is model selection, which involves optimising (or perhaps integrating) model parameters over an objective function.  In the case of Gaussian probabilities, researchers may want to maximise the likelihood $F(\regionA)$ by varying $\{\m,K\}$ subject to some other problem constraints.  Numerical integration methods do not offer this feature currently.  Because the EP approach yields the analytic approximation shown in Equation \ref{eqn:logZ}, we can take derivatives of $\log Z$ so as to approximate $\partial F(\regionA) / \partial \m$ and $\partial F(\regionA) / \partial K$ for use in a gradient solver.  Those derivatives are detailed in Appendix \ref{sec:deriv}.

As a caveat, it is worth noting that numerical integration methods like Genz were likely not designed with these derivatives in mind.  Thus, though it has not been shown in the literature, perhaps an implementation could be derived to give derivative approximations via a clever reuse of integration points (though it is not immediately clear how to do so, as it would be in a simple Monte Carlo scheme).  It is an advantage of the EPMGP algorithm that it can calculate these derivatives naturally and analytically, but we do not claim that it is necessarily a fundamental characteristic of the EP approach unshared by any other approaches.

\subsubsection{Fast runtime and low computational overhead}

By careful consideration of rank one updates and a convenient reparameterisation, we can derive a quadratic run time factor update as opposed to a naive cubic runtime (in the dimensionality of the Gaussian $n$).  This quadratic factor update is done at each iteration, in addition to a determinant calculation in the $\log Z$ calculation, leading to a complexity of $\mathcal{O}(n^3)$ in theory (or $\mathcal{O}(mn^2 + n^3)$ in the case of polyhedral regions).  Numerical integration methods such as Genz also require a Cholesky factorisation of the covariance matrix $K$ and are thus $\mathcal{O}(n^3)$ in theory, and we have not found a more detailed complexity analysis in the literature.  Further, the Genz method uses potentially large numbers of sample points (we use $5\times 10^5$ to derive our high accuracy estimates), which can present a serious runtime and memory burden.    As such, we find in practice that the EP-based algorithm always runs as fast as the Genz methods, and typically runs one to three orders of magnitude faster (the hyperrectangular EPMGP runs significantly faster than the general EPMGP, as the factor updates can be batched for efficiency in languages such as MATLAB).  Furthermore, we find that EP always converges to the same fixed point in a small number of steps (typically 10 or fewer iterations through all factors).    Thus, we claim there are computational advantages to running the EPMGP algorithm over other numerical methods.

There are three caveats to the above claims.  First, since Genz numerical integration methods have a user-selected number of sample points, there is a clear runtime-accuracy tradeoff that does not exist with EP.  EP is a deterministic approximation, and it has no guarantees about convergence to the correct $F(\regionA)$ in the limit of infinite computation.  Claims of computational efficiency must be considered in light of these two different characteristics.  Second, since both EP and Genz algorithms are theoretically $\mathcal{O}(n^3)$, calculated runtime differences are highly platform dependent.  Our implementations use MATLAB, the runtime results of which should not be trusted as reflective of a fundamental feature of the algorithm.  Thirdly, since it is unclear if Genz's method  (available from author) was written with computational efficiency in mind, a more rigorous comparison seems unfair.  Nonetheless, Section \ref{sec:results} presents runtime/accuracy comparisons for a handful of representative cases so that the applied researcher may be further informed about these methods.

Despite these caveats, the EPMGP algorithm runs very fast and stably up to problems of high dimension, and it always converges quickly to what appears to be a unique mode of the EP energy landscape (empirical observation).  Thus, we claim that runtime is an attractive feature of EP for calculating Gaussian probabilities.

\subsubsection{Tail probabilities}

In addition to fast runtime, careful reparameterisation and consideration of the EP parameters leads to a highly numerically stable implementation that is computed in log space.  The lengthy but simple details of these steps are presented in Appendix \ref{sec:epmgpstable}.  Importantly, this stability allows arbitrarily small ``tail" probabilities to be calculated, which are often important in applied contexts.  In our testing the EPMGP algorithm can calculate tail probabilities with $\log Z < -10^5$, that is, $F(\regionA) \approx 10^{-50000}$.  This is vastly below the range of Genz and other methods, which snap to $\log Z = -\infty$ anywhere below roughly $\log Z \approx -150$.  These numbers reflect our own testing of the two methods.  To be able to present results from both methods, we present the results of Section \ref{sec:results} over a much smaller range of probabilities, in the range of $\log Z \approx -20$ to $\log Z = 0$ ($F(\regionA)$ from $10^{-9}$ to $1$).

Again, it is worth noting that the Genz method was likely not designed with tail probabilities in mind, and perhaps an implementation could be created to handle tail probabilities (though it is not obvious how to do so).  Thus, it is an advantage of the EPMGP algorithm that it can calculate tail probabilities naturally, but we do not claim that it is necessarily a unique characteristic of the EP approach.

\subsection{The fundamental similarity between hyperrectangular and polyhedral integration regions}
\label{sec:white}
\begin{figure}
\centering
\hspace{0.0cm}
\includegraphics[width=6in]{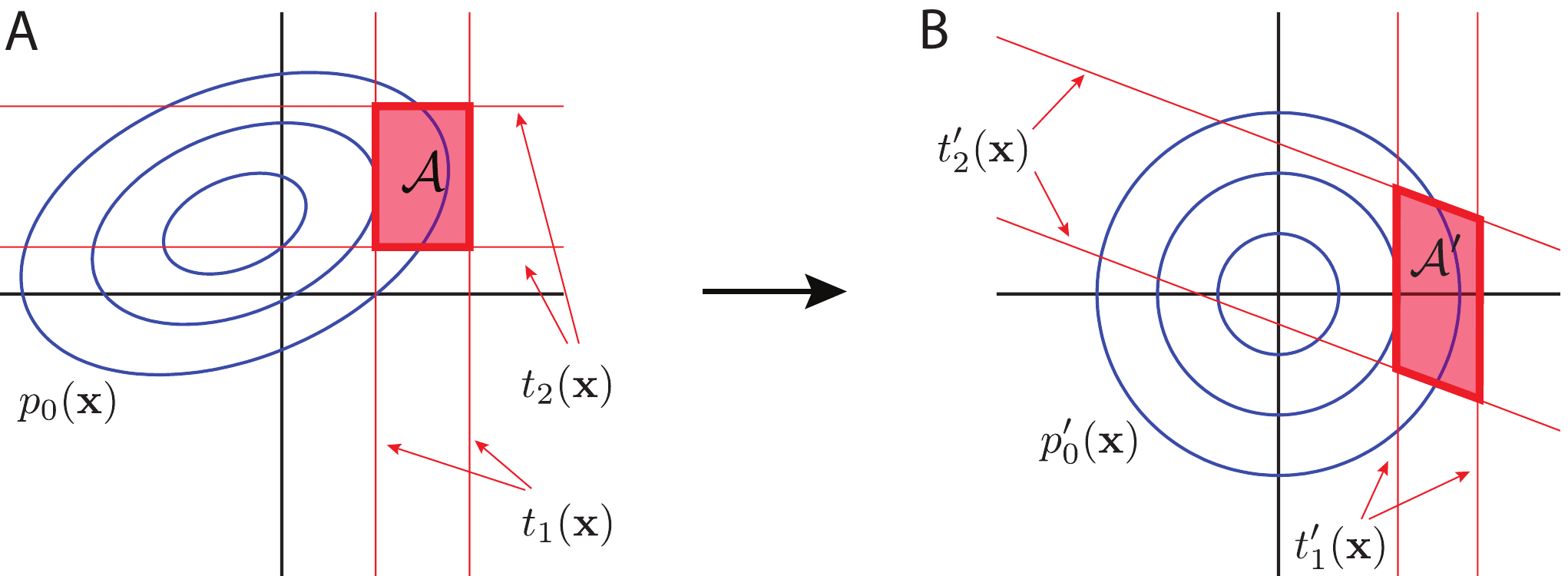}
\caption{\small{The effect of whitening the Gaussian probability based on the covariance of the Gaussian.  Panel A shows a Gaussian probability problem in original form, with the integration region in red.  Panel B shows the same problem when we have whitened (and mean-centered) space with respect to the Gaussian.  We can then think of these problems only in terms of the geometry of the ``coloured" integration region in the whitened space.}}
\label{fig:white} 
\end{figure}

We here note important similarities between the hyperrectangular and polyhedral cases.  Though brief, this section forms the basis for much of the discussion and exploration of the strengths and weaknesses of EP.  Since we have developed EP for arbitrary polyhedral Gaussian probabilities, we can somewhat simplify our understanding of the problem.  We can always whiten the space of the problem via a substitution $\y = L^{-1}(\x - \m)$ where $K = LL^T$.   In this transformed space, we have a standard $\mathcal{N}(0,I)$ Gaussian, and we have a transformed set of polyhedral faces $L^T\c_i$ (which must also be normalised as the $\c_i$ are).   This whitening is demonstrated in Figure \ref{fig:white}.   

Thus, in trying to understand sources of error, we can then attend to properties of this transformed region.  Intuition tells us that, the more hyperrectangular this transformed region is, the better performance, as hyperrectangular cases (in a white space) decompose into a product of solvable univariate problems.  On the other hand, the more redundant/colinear/coloured those faces are, the higher the error that we should expect.    We will thus consider metrics on the geometry of transformed integration region that exists in the whitened space. Further, we then see that hyperrectangular cases are in fact not a fundamental distinction.  Indeed they are simply polyhedra where $m=n$, that is, the number of polyhedral constraints equals the number of dimensions of the problem.  This seemingly small constraint, however, often makes the difference between an algorithm that performs well and one that does not, as the results will show.  

Finally, it is important to connect this change to our EP approach.  Because each step of EP minimises a KL-divergence on distributions, and because KL is invariant to
invertible change of variables, each step of EP is invariant to invertible change of variables.  This invariance requires that the factors, approximate factors, etc. are all transformed accordingly, so that the overall distribution is unchanged (as noted in \cite{seeger08epexpfam}, Section 4). For Gaussian EP, the approximate factors are Gaussian, so the family is invariant to invertible \emph{linear} transformations.  Thus we can equivalently analyse our Gaussian EP algorithm for multivariate Gaussian probabilities in the white space, where $p_0(\x)$ is a spherical Gaussian as in Panel B of Figure \ref{fig:white}.   Considering the whitened space and transformed integration region is conceptually useful and theoretically sound.  This consideration allow us to maintain two views which have different focus: covariance structure in coloured space and a rectangular region (in the hyperectangular case), and geometry of the transformed polyhedron in whitened space.  Much more will be made of this notion in the Results and Discussion.


\section{Other Multivariate Gaussian Probability algorithms}
\label{sec:othermgp}

This section details other methods to which we compare EPMGP.  First, we describe the algorithms due to Genz (Section \ref{sec:Genz}), which represent the state of the art for calculating Gaussian probabilities.  Second, we briefly describe Monte Carlo-based methods (Section \ref{sec:mcmc}).  Third, in some special cases there exist exact ({\it i.e.}, exact to machine precision) methods; we describe those in brief in Section \ref{sec:specialcases}.  EPMGP and the algorithms discussed here will be the algorithms studied in numerical simulation.

\subsection{Genz Numerical Integration Method}
\label{sec:Genz}

The Genz method (first described in \cite{genz92}, but the book \cite{genzBook} is an excellent current reference) is a sophisticated numerical integration
method for calculating Gaussian probabilities.
The core idea is to make a series of transformations of the region $\regionA$ and the Gaussian
$p_0(\x)$ with the hope that this transformed region can be accurately
integrated numerically. This approach performs three
transformations to $F(\regionA)$.  It first 
whitens the Gaussian integrand $p_0(\x)$ (via a Cholesky factorisation of
$K$, which changes the
integration bounds), and secondly it removes the integrand
altogether by transforming the integration bounds with a function
using the cdf and inverse cdf of a
univariate Gaussian.  Finally, a further transformation is done
(using points uniformly drawn from the $[0,1]$ interval) to
set the integration region to the unit box (in $\reals^n$).  Once this is
done, \cite{genz92} makes intelligent choices on dimension ordering
to improve accuracy.  With all orderings and
transformations completed, numerical integration is
carried out over the unit hypercube, using either quasi-random integration
rules \cite[]{nieder72,cranley76} or lattice-point rules
\cite[]{cranley76, nuyens04}.   The original algorithm
\cite[]{genz92} reported, ``it is possible to reliably compute
moderately accurate multivariate normal probabilities for practical
problems with as many as ten variables.'' Further developments
including \cite[]{genz99, genz99b, genz02, genz04} have improved the
algorithm considerably and have generalised it to general polyhedra.
Our use of this algorithm was enabled by the author's MATLAB code
{\tt QSILATMVNV}, which is directly analogous to EPMGP for axis-aligned hyperrectangles.  The author's more general algorithm for arbitrary polyhedra is available from the same code base as {\tt QSCLATMVNV}: this is directly comparable to EPMGP.  These functions run a vectorised,
lattice-point version of the algorithm from \cite[]{genz92}, which we found always
produced lower error estimates (this algorithm approximates the
accuracy of its result to the true probability) than the vectorised, quasi-random
{\tt QSIMVNV} (with similar run-times).  We also found that these vectorised versions
had lower error estimates and significantly faster run-times than
the non-vectorised {\tt QSIMVN}.  Thus, all results shown as ``Genz" were gathered using {\tt QSCLATMVNV}, which was available at the time of this
report at: {\tt
http://www.math.wsu.edu/faculty/genz/software/software.html}.  The
software allows the user to define the number of lattice points used
in the evaluation.  For the Genz methods to which EPMGP was compared, we used $5 \times 10^5$ points.  We
found that many fewer points increased the error
considerably, and many more points increased run-time with little
improvement to the estimates.  We
note also that several statistical software packages have bundled
some version of the Genz method into their product ({\it e.g.}, {\tt mvncdf}
in MATLAB, {\tt pmvnorm} in R).  Particular choices and calculations
outside the literature are typically made in those implementations
(for example, MATLAB does not allow $n>25$), so we have
focused on just the code available from the author himself.

\subsection{Sampling methods}
\label{sec:mcmc}

It is worth briefly noting that Monte Carlo methods are not particularly well suited to this problem.  In the simplest case, one could use a rejection sampler, drawing samples from $p_0(\x)$ and rejecting them if outside the region $\regionA$ (and calculating the fraction of kept samples).  This method requires huge numbers of samples to be accurate, and it scales very poorly with dimension $n$.  The problem with this simple Monte Carlo approach is that the draws from $p_0(\x)$ neglect the location of the integration region $\regionA$, so the rejection sampler can be terribly inefficient.  We also considered importance sampling from the region $\regionA$ (uniformly with a ``pinball" or ``hit-and-run" MCMC algorithm \cite[]{herbrichBook}).  However, this approach has the complementary problem of wasting samples in parts of $\regionA$ where $p_0(\x)$ is small.  Further, this scheme is inappropriate for open polyhedra.  In an attempt to consider both $\regionA$ and $p_0(\x)$, we used elliptical slice sampling \cite[]{murrayESS} to draw samples from $p(\x) = p_0(\x)\prod_i t_i(\x)$ (the familiar Bayesian posterior).  We then calculated the empirical mean and covariance of these samples and used that posterior Gaussian as the basis of an importance sampler with the true distribution.  Though better than naive Monte Carlo, this and other methods are deeply suboptimal in terms of runtime and accuracy, particularly in light of the Genz algorithms, which are highly accurate for calculating the ``true" value, which allows a baseline for comparison to EPGMP.  Thus we will not consider samplers further in this paper.

\subsection{Special case methods}
\label{sec:specialcases}

It is also useful to consider special cases where we can analytically find the true probability $F(\regionA)$.  The most obvious is to consider hyperrectangular integration regions over a diagonal covariance $K$.  Indeed both EPMGP and Genz methods return the correct $F(\regionA)$ to machine precision, since these problems trivially decompose to univariate probabilities.  While a good sanity check, this finding is not instructive.  Instead, here we consider a few cases where the integration region yields a geometric solution in closed form.

\subsubsection{Orthant Probabilities} 

Orthants generalise quadrants to high dimension, for example the nonnegative orthant $\{ \x \in \reals^n : x_i > 0~ \forall i \}$.   For zero-mean Gaussians of low dimension, calculating the probability $F(\regionA)$ when $\regionA$ is an orthant can be solved geometrically.  In $\reals^2$, any orthant (quadrant) can be whitened by the Gaussian covariance $K$, which transforms the orthant into a wedge.  The angle subtended by that wedge, divided by $2\pi$, is the desired probability $F(\regionA)$.  With that intuition in $\reals^2$, researchers have also derived exact geometric solutions in $\reals^3, \reals^4,$ and $\reals^5$ \cite[]{genzBook, sinnKeller2010}.  We can thus compare EP and Genz methods to the true exact answers in the case of these orthant probabilities.  Orthant probabilities also have particularly important application in applied settings such as probit analysis \citep{AshfordSowden1970,SicklesTabuman1986,Gibbons1996}, so knowing the performance of various methods here is valuable.

\subsubsection{Trapezoid Probabilities}

Trapezoid probabilities are another convenient construction where we can know the true answer.   If we choose our Gaussian to be zero-mean with a diagonal covariance K, and we choose $\regionA$ to be a hyperrectangle that is symmetric about zero (and thus the centroid is at zero), then this probability decomposes simply into a product of univariate probabilities.  We can then take a non-axis cut through the center of this region.   If we restrict this cut to be in only two dimensions, then the region in those two dimensions is now a trapezoid (or a triangle if the cut happens to pass through the corner, but this happens with zero probability).   By the symmetry of the Gaussian and this region, we have also cut the probability in half.  Thus, we can calculate the simple rectangular probability and half it to calculate the probability of this generalised trapezoid.   

The virtue of this construction is that this trapezoidal region appears as a nontrivial, general polyhedron to the EPMGP and Genz methods, and thus we can use this simple example to benchmark those methods.  Unlike orthant probabilities, trapezoids are a contrived example so that we can test these methods.  We do not know of any particular application of these probabilities, so these results should be seen solely for vetting of the EP and Genz algorithms.

\section{Cases for Comparing Gaussian Probability Algorithms}
\label{sec:testcases}

The following section details our procedure for drawing example integration regions $\regionA$ and example Gaussians $p_0(\x)$ for the numerical experiments.  Certainly understanding the behavior of these algorithms across the infinite space of parameter choices is not possible, but our testing in several different scenarios indicates that the choices below are representative.

\subsection{Drawing Gaussian distributions $p_0(\x) = \mathcal{N}(\x; \m,K)$}

To define a Gaussian $p_0(\x)$, we first randomly
drew one positive-valued $n$-vector from an exponential distribution with mean
10 ($\lambda = 0.1$, $n$ independent draws from this distribution),
and we call the corresponding diagonal matrix the matrix of eigenvalues $S$.  We then
randomly draw an $n\times n$ 
orthogonal matrix $U$ (orthogonalising a matrix of iid $\mathcal{N}(0,1)$ random variables with the singular
value decomposition suffices), and we form the Gaussian covariance matrix
$K = USU^T$.  This procedure produces a good spread of differing
covariances $K$ with quite different eigenvalue 
spectra and condition numbers. We
note that this procedure produced a more interesting range of
$K$ - in particular, a better spread of condition numbers - than using draws from a Wishart distribution.  We
then set the mean of the Gaussian $\m=0$ without loss of
generality (note that, were $\m\neq 0$, we could equivalently
pick the region $\regionA$ to be shifted by $\m$, and then the
problem would be unchanged; instead, we leave $\m=0$, and we
allow the randomness of $\regionA$ to suffice).   For orthant probabilities, no change to this procedure is required.  For trapezoid probabilities, a diagonal covariance is required, and thus we set $U = I$ in the above procedure.

\subsection{Drawing integration regions $\regionA$}

Now that we have the Gaussian $p_0(\x) = \mathcal{N}(\x; \m,K)$,
we must define the region $\regionA$ for the Gaussian probability
$F(\regionA)$.  

\subsubsection{Axis-aligned hyperrectangular regions}  

The hyperrectangle $\regionA$ can be defined by two
$n$-vectors: the upper and lower bounds $\u$ and $\l$, where each entry corresponds to a dimension of the problem.  To make
these vectors, we first randomly drew a point from the Gaussian
$p_0(\x)$ and defined this point as an interior point of
$\regionA$.  We then added and subtracted randomly chosen lengths to
each dimension of this point to form the bounds $u_i$ and $l_i$.  Each length from the interior point was chosen uniformly between 0.01 and the dimensionality $n$.  Having the region size scale with the dimensionality $n$ helped to prevent the probabilities $F(\regionA)$ from becoming vanishingly small with increasing dimension (which, as noted previously, is handled fine by EP but problematic for other methods).  

\subsubsection{Arbitrary polyhedral regions}

The same procedure is repeated for drawing the interior point and the bounds $l_i$ and $u_i$ for all $m$ constraints.  The only difference is that we also choose axes of orientation, which are unit-norm vectors $\c_i \in \reals^n$ for each of the $m$ factors.  These vectors are drawn from a uniform density on the unit hypersphere (which is achieved by normalising $n$ iid $\mathcal{N}(0,1)$ draws).  The upper and lower bounds are defined with respect to these axes.

With these parameters defined, we now have a randomly
chosen $\regionA$ and the Gaussian $p_0(\x)$, and we can test all methods as previously described.  This
procedure yields a variety of Gaussians and regions, so we believe our results
are strongly representative. 

It is worth noting here that arbitrary polyhedra are not uniquely defined by the ``intersection of boxes" or more conventional ``intersection of halfspaces" definition.  Indeed, there can be inactive, redundant constraints which change the representation but not the polyhedron itself.  We can define a minimal representation of any polyhedron $\regionA$ to be the polyhedron where all constraints are active and unique.  This can be achieved by pulling in all inactive constraints until they support the polyhedron $\regionA$ and are thus active, and then we can prune all repeated/redundant constraints.  These steps are solvable in polynomial time as a series of linear programs, and they bear important connection to problems considered in centering (analytical centering, Chebyshev centering, etc.) for cutting-plane methods and other techniques in convex optimisation ({\it e.g.}, see \cite{boydBook}).  These steps are detailed in Appendix \ref{sec:minpoly}.   To avoid adding unnecessary complexity, we do not minimalise polyhedra in our results, as that can mask other features of EP that we are interested in exploring.  Anecdotally, we report that while minimalising polyhedra \emph{will} change EP's answer (a different set of factors), the effect is not particularly large in these randomly sampled polyhedral cases.  In our empirical and theoretical discussion of the shortcomings of EP, we will heavily revisit this notion of constraint redundancy.

\section{Results}
\label{sec:results}

We have thus far claimed that EP provides a nice framework for Gaussian probability calculations, and it offers a number of computational and analytical benefits (runtime, tail probabilities, derivatives).   We have further claimed that existing numerical integration methods, of which the Genz method is state of the art, can reliably compute Gaussian probabilities to high accuracy, such that they can be used as ``ground truth" when evaluating EP.  Thirdly, we claim that this Gaussian probability calculation elucidates a pathology in EP that is not often discussed in the literature.  Giving evidence to those claims is the purpose of this Results section, which will then be discussed from a theoretical perspective in Section \ref{sec:discussion}.

The remainder of this results section is as follows:  we first demonstrate the accuracy of the EP method in the hyperrectangular case across a range of dimensionalities (Section \ref{sec:axisresults}).  Using the same setup, we then demonstrate a higher error rate for the general polyhedral case in Section \ref{sec:generalresults}, testing the sensitivity of this error to both dimensionality $n$  and the number of polyhedral constraints $m$.  In Section \ref{sec:specialresults}, we evaluate the empirical results for the special cases where the true Gaussian probability can be calculated by geometric means, as introduced in the previous Section \ref{sec:specialcases}, which allows us to compare the EP and Genz approaches.  We next test the empirical results for the commonly tested case of equicorrelated Gaussian probabiliites (Section \ref{sec:equicorrresults}).  The equicorrelation and special case results point out the shortcomings of the EP approach.  We can then proceed to construct pathological cases that clarify where and why EP gives poor approximations of Gaussian probabilities, which is done in Section \ref{sec:pathologicalresults}.  This final section gives the reader guidelines about when EP can be effectively used for Gaussian probabilities, and it provides a unique perspective and leads to a theoretical discussion about negative aspects of EP rarely discussed in the literature.

\subsection{EP results for hyperrectangular integration regions}
\label{sec:axisresults}

\begin{figure}
\centering
\hspace{0.0cm}
\includegraphics[width=6in]{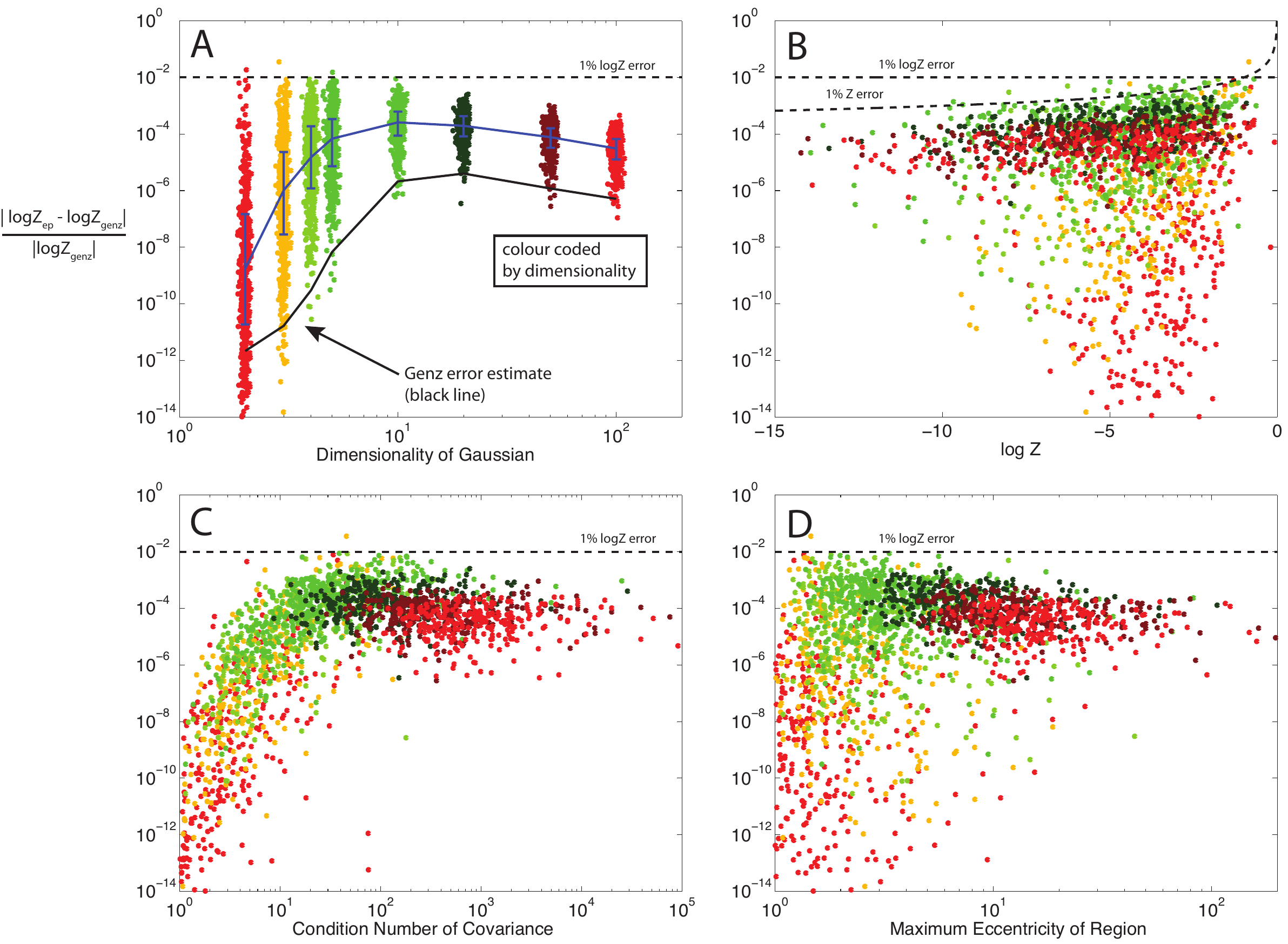}
\caption{\small{Empirical results for EPMGP with hyperrectangular integration regions.  Each point represents a random Gaussian $p_0(\x)$ and a random region $\regionA$, as described in Section \ref{sec:testcases}.  In all panels the colour coding corresponds to the dimensionality of the Gaussian integral (chosen to be $n = \{2,3,4,5,10,20,50,100\}$.  All panels show the relative log error of the EP method with respect to the high accuracy Genz method.  Panel A plots the errors of EPMGP by dimensionality (note log-log scale).  The horizontal offset at each dimension is added jitter to aid the reader in seeing the distribution of points (panel A only).  The blue line and error bars plot the median and $\{25\%,75\%\}$ quartiles at each dimension $n$.  The black dashed line indicates the $1\% \log Z$ error.  Finally, the black line below the EP results is the median error estimate given by Genz.  Since this line is almost always a few orders of magnitude below the EP distribution, it is safe to use the high accuracy Genz method as a proxy to ground truth.  Panel B plots the errors by the value of $\log Z$ itself.  On this panel we also plot a $1\%$ error for $Z\approx F(\regionA)$.  Panel C repeats the same errors but plotted by the condition number of the covariance $K$.  Panel D plots the errors vs the maximum eccentricity of the hyperrectangle, namely the largest width divided by the smallest width.}}
\label{fig:axisresults} 
\end{figure}

Figure \ref{fig:axisresults} shows the empirical results for the EPMGP algorithm in the case of hyperrectangular integration regions.  For each dimension $n = \{2,3,4,5,10,20,50,100\}$, we used the
above procedure to choose a random Gaussian $p_0(\x)$ and a random region $\regionA$.   This procedure was repeated 1000 times at each $n$.  Each choice of $p_0(\x)$ and $\regionA$ defines a single Gaussian probability problem, and we can use EPMGP and Genz methods to approximate the answer.  We calculate the relative error between these two estimates.  Each of these errors is plotted as a point in Figure \ref{fig:axisresults}.  We plot only 250 of the 1000 points at each dimensionality, for purposes of visual clarity in the figures.  The four panels of Figure \ref{fig:axisresults} plot the same data on different $x$-axes, in an effort to demonstrate what aspects of the problem imply different errors.  As described in the figure caption, Panel A plots the errors of EPMGP by dimensionality $n$.  Horizontal offset at each dimension is added jitter to aid the reader in seeing the distribution of points (panel A only).  The black dashed line indicates the $1\% \log Z$ error.  Finally, the black line below the EP results is the median error estimate given by Genz.  Since this line is almost always a few orders of magnitude below the distribution of EP errors, it is safe to use the high accuracy Genz method as a proxy to ground truth.   

Panel A demonstrates that the EP method gives a reliable estimate, typically with errors on the order of $10^{-4}$, of Gaussian probabilities with hyperrectangular integration regions.  The shape of the blue curve in Panel A seems to indicate that error peaks around $n=10$.  We caution the reader not to over-interpret this result, as this feature may be just as much about the random cases chosen (as described in Section \ref{sec:testcases}) as a fundamental feature of Gaussians.  Nonetheless, we do notice across many test cases that error decreases at higher dimension, perhaps in part due to the concentration of mass of high dimensional Gaussians on the $1\sigma$ covariance ellipsoid.

Panel B plots the errors by the value of $\log Z$ itself, where we use the $\log Z$ calculated from Genz.  On this panel we can also plot a $1\%$ error for $Z\approx F(\regionA)$, as the log scale can suggest misleadingly large or small errors.  A priori we would not expect any significant trend to be caused by the actual value of the probability, and indeed there seems to be a weak or nonexistent correlation of error with $\log Z$.  Panel C repeats the same errors but plotted by the condition number of the covariance $K$.   In this case perhaps we would expect to see trend effects.  Certainly the EP algorithm should be correct to machine precision in the case of a hyperrectangular region and a white covariance, since this problem decomposes into $n$ one-dimensional problems.  Indeed we do see this accuracy in our tests.  Increasing the eccentricity of the covariance, which can be roughly measured by the condition number, should and does increase the error, as shown in Panel C.  This effect can be seen more clearly in later figures (when we test equicorrelation matrices).  One might also wonder about the condition number of the correlation, as this value would neglect axis-scaling effects.  We have also tested this and do not show this panel as it looks nearly identical to Panel C.  

Panel D plots the errors vs the maximum eccentricity of the hyperrectangle, namely the largest width divided by the smallest width.  As with Panel B, it is unclear why the eccentricity of a hyperrectangular region should have bearing on the accuracy of the computation, and indeed there appears to be little effect.  The colour coding helps to clarify this lack of effect, because each dimensionality seems to have insensitivity to error based on changing eccentricity.  Thus any overall trend appears more due to the underlying explanatory factor of dimensionality than due to the region eccentricity itself.  

Finally, as an implementation detail, we discuss the Genz error estimate (as seen in the black line of Panel A).  It is indeed true that this error estimate can sometimes be significant, but that happens rarely enough ($<10\%$) that it has little bearing on the trends shown here.  To be strict, one can say that the lowest $10\%$ of EP errors are conservative, noisy errors: they could be significantly smaller, but not significantly larger.  Due to the log scale of these plots, the larger EP errors (the ones we care about) are reliable.  Furthermore, we note that the Genz errors increase significantly at higher dimension $n>100$.  The EP method can compute larger dimensions without trouble, but the Genz method achieves a high enough error estimate that many more, often a majority, of EP errors are to be considered conservative, noisy estimates.  Empirically, it seems the errors in higher dimensions continue their downward trend, but they are noisier measurements, so we do not report them to avoid over penalising the EP method (as they could be much smaller).

Overall, the empirical results shown here indicate that EPMGP is a successful candidate algorithm for Gaussian probabilities over hyperrectangular integration regions.  The error rate is non-zero but generally quite low, with median errors less $10^{-4}$ and individual errors rarely in excess of 1\% across a range of dimensions, which may be acceptable in many applied settings.

\subsection{EP results for general polyhedral cases}
\label{sec:generalresults}

\begin{figure}
\centering
\hspace{0.0cm}
\includegraphics[width=6in]{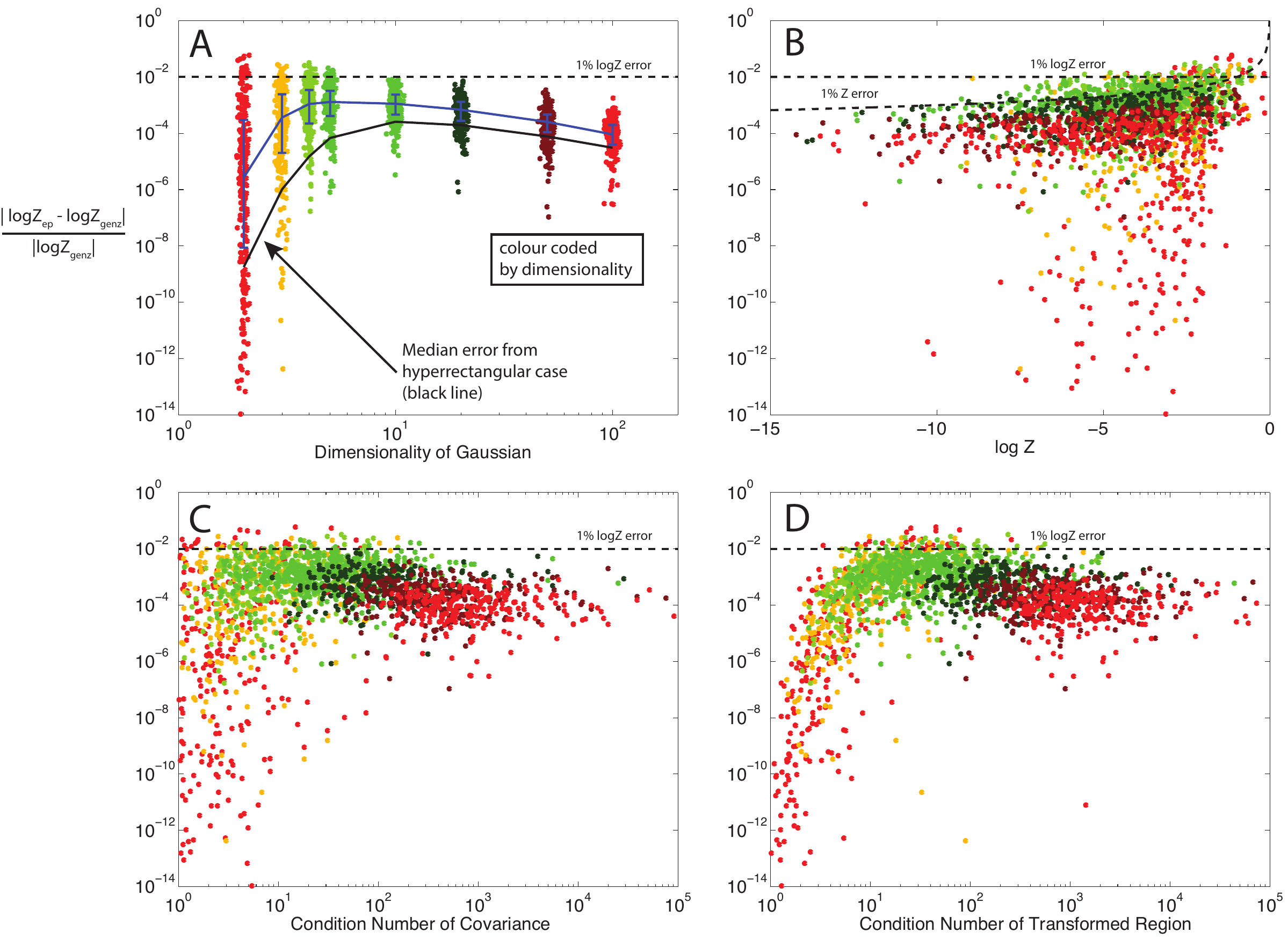}
\caption{\small{Empirical results for EPMGP with polyhedral integration regions.  This figure has largely the same description as Figure \ref{fig:axisresults}.  The only difference from that previous figure is that panel A depicts as a black line the median errors from EPMGP in the hyperrectangular case.  This demonstrates the consistently higher errors of general polyhedral EPMGP.  Panels B and C have the same description as Figure \ref{fig:axisresults}.  Panel D plots the condition number of the transformed region.  As discussed in the text, any Gaussian probability problem can be whitened to consider a Gaussian with identity covariance.  This step merely linearly transforms the region $\regionA$.  Panel D shows the errors plotted vs. the condition number of the integration region transformed appropriately by the Gaussian covariance $K$.}}
\label{fig:generalresults} 
\end{figure}
We have previously shown how the EP framework readily can be extended to arbitrary polyhedral regions with only a minor change to the calculation of cavity parameters (Equation \ref{eqn:cavityepmgpPoly}).  Figure \ref{fig:generalresults} shows the empirical results of this change, which are perhaps surprisingly large.  In the same way we drew 1000 test cases at each dimensionality $n$.  The only difference is that we also drew $m = n$ random polyhedral faces as described in Section \ref{sec:testcases}, instead of the fixed hyperrectangular faces in the previous case.  

Figure \ref{fig:generalresults} has the same setup as \ref{fig:axisresults}, with only a few exceptions.  First, the black line in Panel A shows the median EP errors in the axis aligned case ({\it i.e.}, the blue line in Figure \ref{fig:axisresults}, Panel A).  The purpose of this line is to show that indeed for the same Gaussian cases, the error rate of polyhedral EPMGP is typically an order of magnitude or two higher than in the hyperrectangular case.  Genz errors are not shown (as they were in Figure \ref{fig:axisresults}), but we note that they tell the same story as in the hyperrectangular case.  The only other change is Panel D.  In Section \ref{sec:white}, we introduced the concept of whitening the space to consider just the region integrated over a standard Gaussian.   We consider this region as a matrix with columns equal to $L^T\c_i$ (normalised), the transformed polyhedral faces.  If this region were hyerrectangular (in the transformed space), then this matrix would be $I$, and the method would be perfectly accurate (decomposed problem).  Thus, to test the transformed region's proximity to $I$, we can use the condition number.  We do so in Panel D, where we plot errors against that value.  As in Panel C, there appears to be some legitimate trend based on the condition number of the transformed region.  This finding confirms our intuition that less hyperrectangular transformed integration regions will have higher error.  There are a number of other metrics on the transformed region that one might consider that attempt to account for colinearity (non-orthogonality) in the transformed region constraints.  We have produced similar plots with Frobenius ($l_2$) and $l_1$ norms of the elements of the scaled Gram matrix (instead of condition number), and we report that the trends found there are similar to these figures: there is some upward trend in error with a metric of colinearity.
\begin{figure}
\centering
\hspace{0.0cm}
\includegraphics[width=6in]{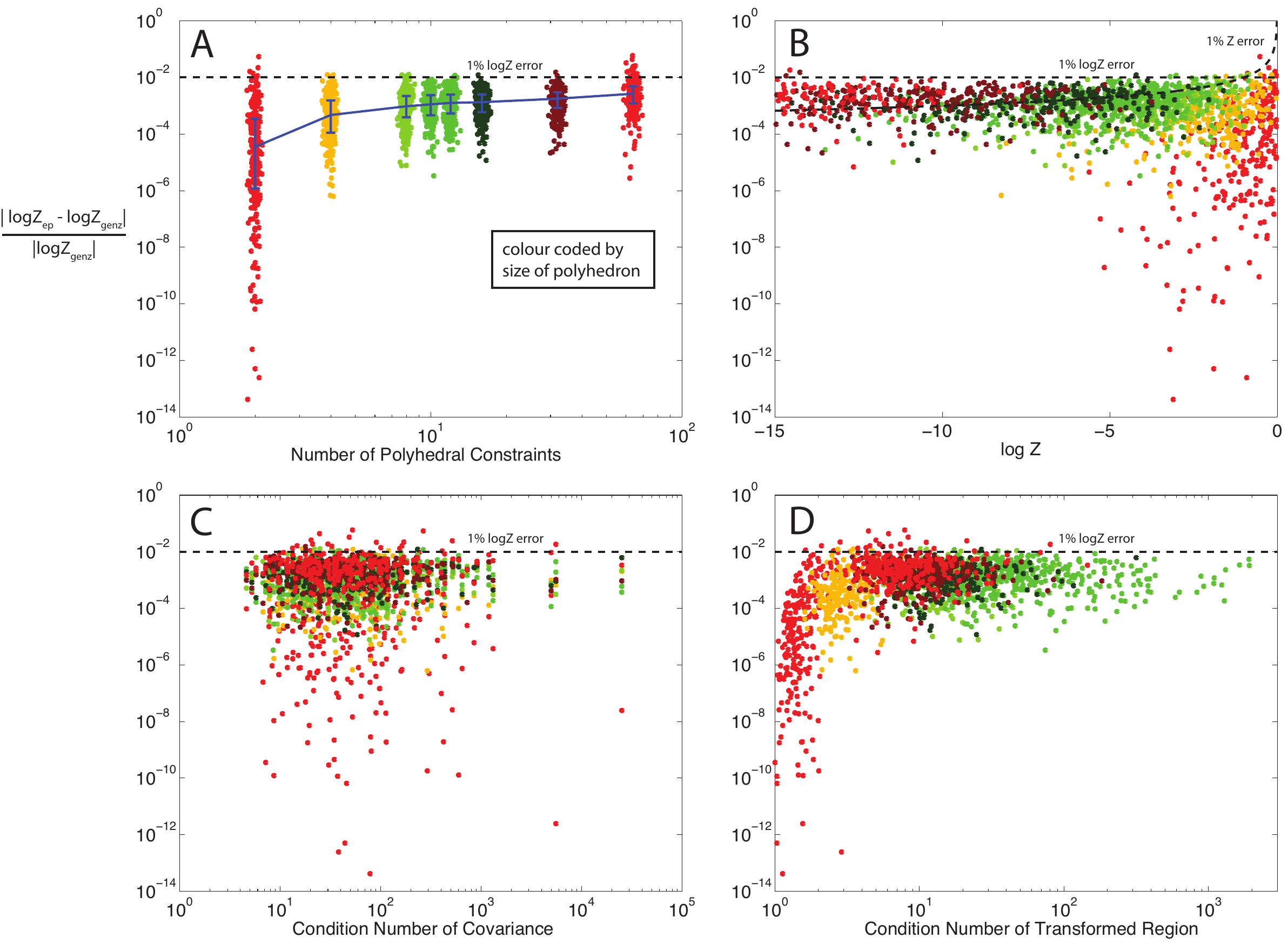}
\caption{\small{Empirical results for EPMGP with polyhedral integration regions.  Panel A calculates the same error metric as in the previous figures for $n=10$ dimensional Gaussians with varying numbers of polyhedral constraints $m = \{2,4,8,10,12,16,32,64\}$.  Note the difference with Figure \ref{fig:generalresults} where dimensionality $n$ is swept (and $m=n$).  Other panels are as previously described in Figure \ref{fig:generalresults}.}}
\label{fig:generalresultspoly} 
\end{figure}

Figure \ref{fig:generalresults} only shows cases where the number of polyhedral constraints is the same as the dimensionality of the Gaussian ($m=n$).  The sensitivity of error to the number of constraints $m$ is also an important question.  Figure \ref{fig:generalresultspoly} gives these empirical results.  We use the same setup as Figure \ref{fig:generalresults} with only one change: Panel A plots errors by number of polyhedral face constraints $m$ instead of by Gaussian dimension $n$.  In this figure all cases use $n=10$ dimensions, and we show polyhedral sizes of $m = \{2,4,8,10,12,16,32,64\}$.  It seems from Panel A that larger numbers of constraints/polyhedral faces do imply larger error, trending up by roughly an order of magnitude or two.  This is an interesting finding that will be revisited in the theoretical discussion: for a given dimension $n$, more polyhedral constraints (more factors) implies higher EP error.  The errors shown in B and C seem largely invariant to the condition number of the covariance and the value $\log Z$ itself.  However, Panel D still shows the error in the whitened space and indicates again some upward trend, as we would expect.

Figure \ref{fig:generalresults} and \ref{fig:generalresultspoly} show error rates that may in many cases be unacceptable.  Though EP still performs well often, the reliability of the algorithm is not what it is in the hyperrectangular case of Figure \ref{fig:axisresults}.   We now test special cases where true answers are known, which will further our understanding of the strengths and weaknesses of the EP approach.

\subsection{EP and Genz results for special cases with known answers}
\label{sec:specialresults}

\begin{figure}
\centering
\hspace{0.0cm}
\includegraphics[width=6in]{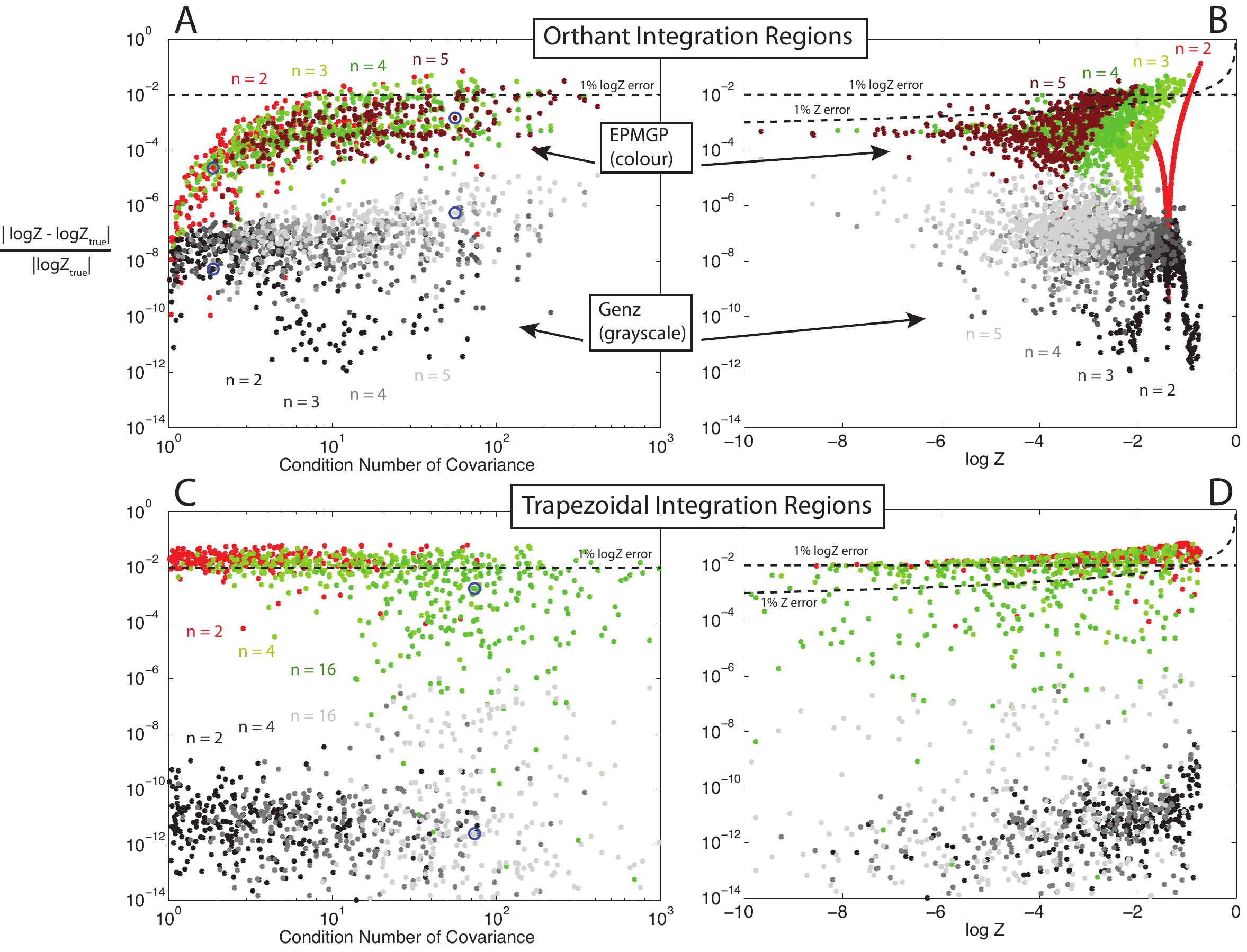}
\caption{\small{Empirical results for the special cases where an analytical true answer can be calculated.  Panels A and B show results for orthant probabilities of dimension $n = \{2,3,4,5\}$ (See Section \ref{sec:specialcases}).   Panel A shows relative $\log Z$ error, and Panel B shows relative $Z$ error.  The coloured points indicate EP errors, and the grayscale points indicate the errors of the Genz method.  Though not shown in this figure, the Genz estimates of error only slightly and occasionally overlap the coloured distribution of EP errors, which again clarifies that Genz can be effectively used as a proxy to ground truth. Panels C and D plot errors for trapezoid probabilities (see Section \ref{sec:specialcases}), again with EP and Genz errors.  Note also the three pairs of points in Panel A and C which are circled in blue.  These are three representative examples: one each of an orthant probability in $n=2$, an orthant probability in $n=5$, and a trapezoid probability in $n=16$.  We use these representative examples in the following Figure \ref{fig:runtimeresults} to evaluate the runtime/accuracy tradeoff of the EP and Genz methods.}}
\label{fig:specialresults} 
\end{figure}

Thus far we have relied on the Genz method to provide the baseline of comparisons.  Though we have argued that this step is valid, it is still useful to consider the handful of special cases where the true probability can be calculated from geometric arguments.  We do so here, and those results are shown in Figure \ref{fig:specialresults}.

Panels A and B of Figure \ref{fig:specialresults}  show errors for both the EP method (colour) and the Genz method (grayscale).  The two panels are the usual errors plotted as a function of condition number of covariance (Panel A) and the true probability $\log Z$ (Panel B).   The four cases shown are orthant probabilities at $n = \{2,3,4,5\}$.  There are a few interesting features worthy of mention.  First, we note that there is a clear separation between the Genz errors and the EP errors (in log-scale no less).  This finding helps to solidify the earlier claim that the Genz numerical answer can be used as a reasonable proxy to ground truth.  We can also view this figure with the Genz error estimates as error bars on the grayscale points.  We have done so and report that the majority of the coloured points remain above those error bars, and thus the error estimate reinforces this difference between Genz and EP.  We do not show those error bars for clarity of the figure.

Second, it is interesting to note that there is significant structure in the EP errors with orthant probabilities when plotted by $\log Z$ in Panel B.  Each dimension has a ``V" shape of errors.  This can be readily explained by reviewing what an orthant probability is.  For a zero-mean Gaussian, an orthant probability in $\reals^2$ is simply the mass of one of the quadrants.  If that quadrant has a mass of $\log Z = \log 0.25$, then the correlation must be white, and hence EP will produce a highly accurate answer (the problem decomposes).  Moving away from white correlation, EP will produce error.  This describes the shape of the red curve for $n=2$, which indeed is minimised at $\log 0.25 \approx -1.39$.  The same argument can be extended for why there should be a minimum in $\reals^3$ at $\log Z = \log 0.125$, and so on.  It is interesting to note that the Genz errors have structure of their own in this view, which is presumably related to the same fact about orthant probabilities.  Investigating the pattern of Genz errors further is beyond the scope of this work, however. 

Panels C and D plot the same figures as Panels A and B, but they do so with trapezoid probabilities.   As described above, trapezoid probabilities involve mean-centered hyperrectangular integration regions and Gaussians with diagonal covariance.  There is then an additional polyhedral constraint that cuts two dimensions of the region through the origin such that the probability is halved.  This appears to EP and to Genz as an arbitrary polyhedron in any dimensionality.  We show here results for $n = \{2,4,16\}$.  Panels C and D clarify that EP works particularly poorly in this case, often with significant errors.  This finding highlights again that EP with general polyhedral regions can yield unreliable results.  
\begin{figure}
\centering
\hspace{0.0cm}
\includegraphics[width=6in]{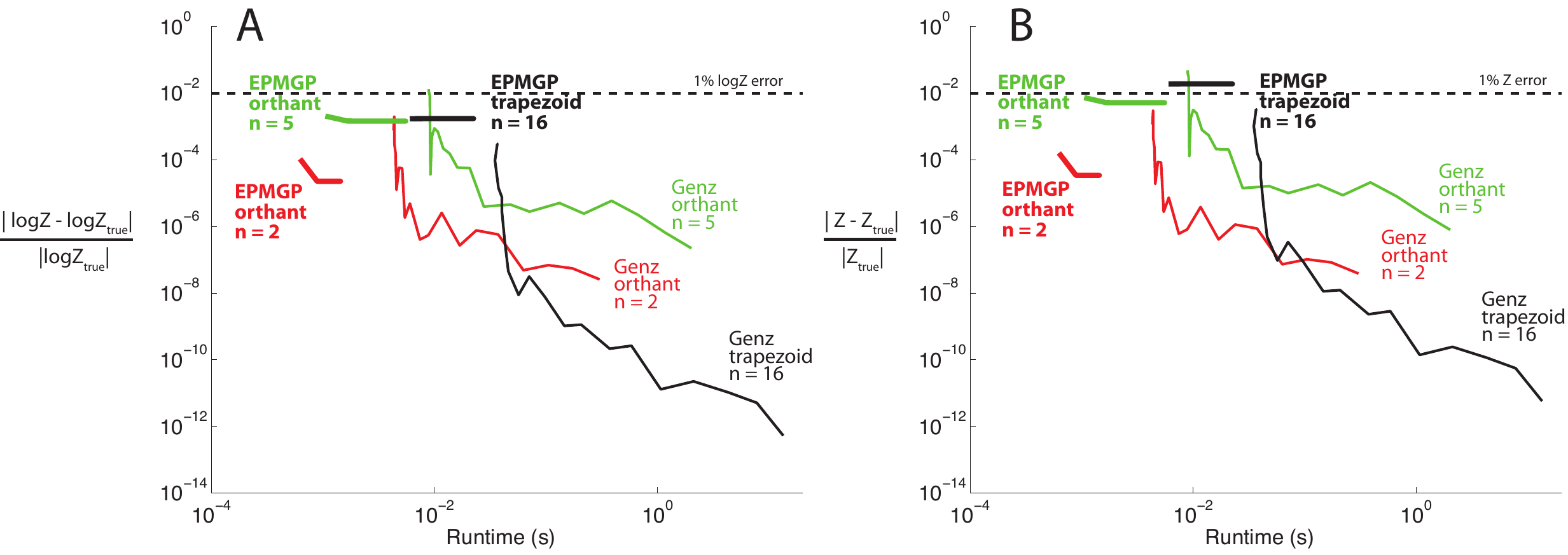}
\caption{\small{Runtime analysis of the EPMGP and Genz algorithms.   The three pairs of lines are three representative examples, which were circled in blue in Figure \ref{fig:specialresults}.  They include orthant probabilities at dimension $n=2$ and $n=5$, and a trapezoid probability at $n=16$.  Each point in this Figure is the median run time of 100 repeated trials (both methods give the same error on each repetition).  The Genz method has a clear runtime/accuracy tradeoff, as more integration points significantly increases accuracy and computational burden.  For Genz we swept the number of integration points from 50 to $10^6$.  The EP method was run with a fixed maximum number of iterations (from 1 to 20).  Because the method converges quickly, rarely in more than 10 steps, these lines typically appear flat.  Note that EP runs one to three orders of magnitude more quickly than Genz, but more computation will not make EP more accurate.  }}
\label{fig:runtimeresults} 
\end{figure}

In Figure \ref{fig:specialresults}, Panels A and C also show 3 pairs of points circled in blue.  Those points are three representative test cases - an orthant probability in $n=2$, an orthant probability in $n=5$, and a trapezoid probability in $n=16$.  We chose these as representative cases to demonstrate runtime/accuracy tradeoffs, which is shown in Figure \ref{fig:runtimeresults}.   This figure is offered for the applied researcher who may have runtime/accuracy constraints that need to be balanced.  Panel A shows the familiar $\log Z$ errors, and Panel B shows the same cases with $Z$ error.  Each colour gives the error plotted as a function of runtime for one of the pair of points circled in Figure \ref{fig:specialresults}.  For the Genz method, the lines are formed by sweeping the number of integration points used from 50 to $10^6$.  For the EP method (results in bold), we vary the number of EP iterations allowed, from 2 to 20.  Since the EP method rarely takes more than 10 iterations to converge numerically, the bold lines are rather flat.   Each point in these panels is the median runtime of 100 repeated runs (the results and errors were identical for each case, as both methods are deterministic).

Figure \ref{fig:runtimeresults} shows that EPMGP runs one to three orders of magnitude faster than the Genz method.  However, since EP is a Gaussian approximation and not a numerical computation of the correct answer, one can not simply run EPMGP longer to produce a better answer.  The Genz method on the other hand has this clear runtime/accuracy tradeoff.   Our results throughout the other figures use the high accuracy Genz result ($5\times 10^5$ integration points - the rightmost point on the Genz lines) and the fully converged EP method (again the rightmost points on the EP lines).  Finally, we note that we chose three representative examples, but these are highly reflective of the typical runtime/accuracy tradeoff we have seen across many more test cases.

\subsection{EP results for equicorrelated Gaussian probabilities}
\label{sec:equicorrresults}

Another interesting case for study is the case of Gaussians with equicorrelated covariance matrices.  These Gaussians have covariances with all diagonal entries equal to 1, and all off-diagonal entries equal to $\rho$, where $-1/(n-1) \le \rho \le 1$ (these bounds ensure a positive semidefinite matrix).    This case is interesting primarily because it allows us to look at a parameterised family of covariance matrices that include the identity matrix ($\rho = 0$, where we believe EP should produce a highly accurate answer) and highly elongated matrices ($\rho \rightarrow 1$ or $\rho \rightarrow -1/(n-1)$, where Figures \ref{fig:generalresults} and \ref{fig:specialresults} suggest that EP should do poorly).  This case is also important as it connects to previous literature on the subject of Gaussian probabilities, where equicorrelated matrices are a common test case (such as \cite{gassmann2002, shervish1984}).

\begin{figure}
\centering
\hspace{0.0cm}
\includegraphics[width=6in]{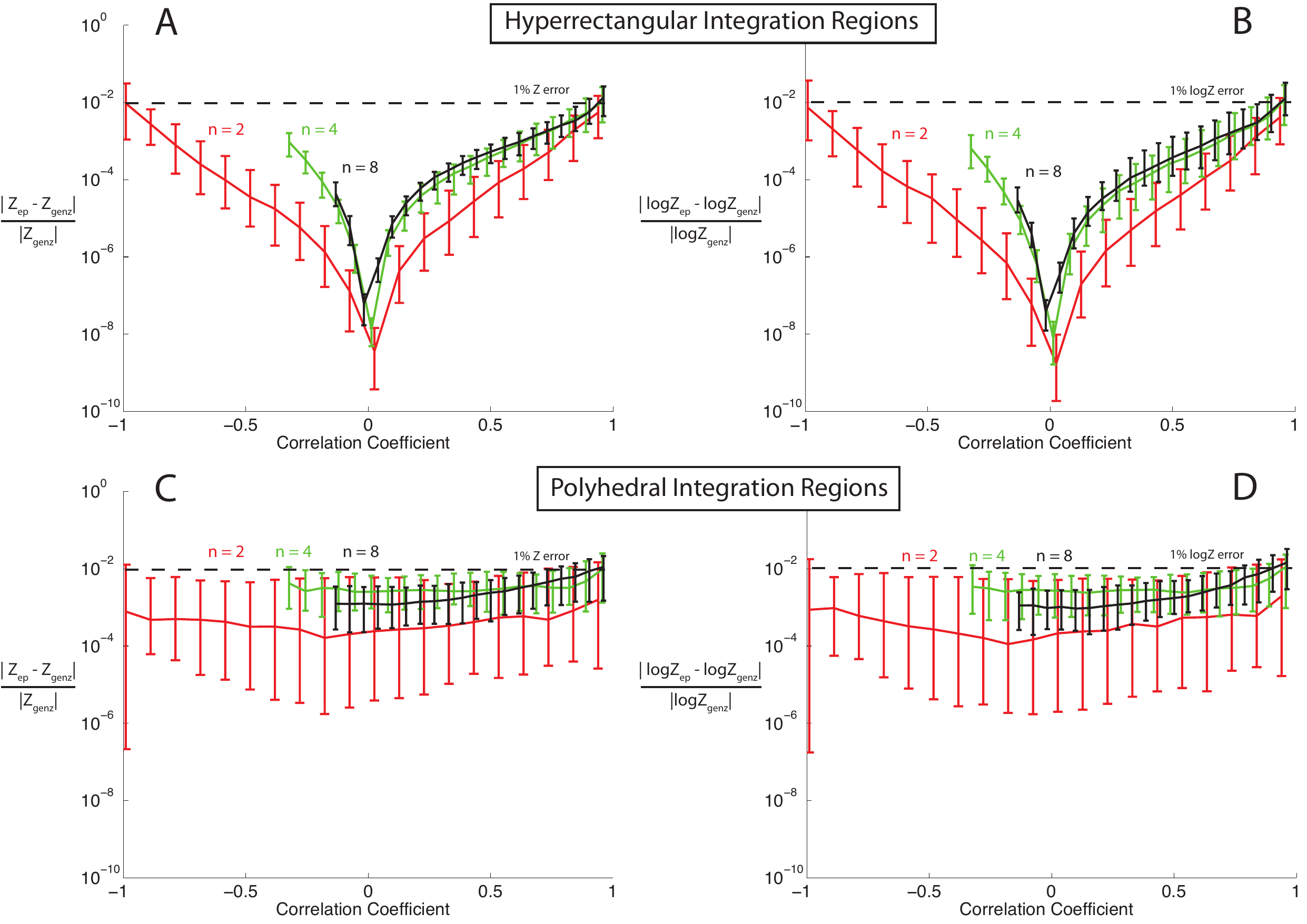}
\caption{\small{ Empirical results for EPMGP over equicorrelated Gaussians.  Historically, a common test case for Gaussian probability algorithms has been to test error on correlation matrices of equal correlation (see {\it e.g.}, \cite{gassmann2002, shervish1984}).  All other quantities - the region $\regionA$ and Gaussian mean $\m$ - are still drawn randomly as previously described.  Here we test error for both the hyperrectangular case (panels A and B) and the polyhedral case (panels C and D), showing relative error in $Z$ (panels A and C) and relative error in $\log Z$ (panels B and D).  We show three different dimensionalities $n = \{2,4,8\}$ coloured as red, green, and black, respectively.  100 random cases were run at each correlation value.  Each error bar shows the median and $\{25\%,75\%\}$ quartiles.  The $x$-axis sweeps the correlation coefficient $\rho$ throughout its range for each dimension (which is the $[-1/(n-1) , 1]$ interval; matrices outside this range are not valid, positive semidefinite correlation matrices). }}
\label{fig:resultsequicorr} 
\end{figure}

Each panel tests the three cases of $n = \{2,4,8\}$, with the lines colour coded as in previous figures.  The $x$-axis of these figures sweeps the valid range of $\rho$ values.  At each $\rho$ value, which fully specifies a Gaussian covariance, we draw the region $\regionA$ and Gaussian mean $\m$ in the usual way, as previously described.  We do this 100 times at each value of $\rho$, calculating the EP and Genz results, from which we can compute errors in $\log Z$ and $Z$ in the usual way.  Again the line and error bars at each point show the median and $\{25\%,75\%\}$ quartiles.   Panels A and B show these results with hyperrectangular integration regions, and Panels C and D show the results with general polyhedral regions.

Panels A and B behave as intuition suggests: when the covariance $K$ is nearly white, the error is minimal.  As the correlation becomes larger and the matrix more eccentric, error increases.  Also in the hyperrectangular case we see that errors become meaningful - around $1\%$ - only in the most eccentric cases where the matrix is nearly singular. 

Panels C and D appear to have a much flatter error, which perhaps disagrees with intuition.  The error seems relatively insensitive to the eccentricity of the correlation structure.  Errors do have a slight downward trend towards 0, but it is by no means significant.  This finding can be readily explained by returning to the notion of the whitened space (Figure \ref{fig:white} and Panel D of Figures \ref{fig:generalresults} and \ref{fig:generalresultspoly}).  This result also helps us transition into understanding pathologies of EP.  Recall that the problem can be whitened by the covariance of the Gaussian, such that the Gaussian is standardised to identity covariance and the region carries all notions of eccentricity.  If we revisit Panels A and B in this light, we see that the EP method is doing well when the transformed region is close to hyperrectangular ($\rho \approx 0$), but then error grows as the region becomes coloured/non-rectangular ($\rho \rightarrow 1$ or $\rho \rightarrow -1/(n-1)$).   The intuition and the results are unchanged in Panels A and B.  Turning to Panels C and D, we can imagine the same effect as $\rho$ is changed.  However, because we are in the case of polyhedral integration regions, the region we draw is \emph{already} coloured before we whiten the space.  Thus, when we whiten the space, we will equally often make a region less hyperrectangular and more hyperrectangular.  In this view, sweeping this correlation matrix should have minimal average effect on the error.  This lack of effect is indeed what we see in Panels C and D of Figure \ref{fig:resultsequicorr}.  

\subsection{Contrived, pathological cases to illustrate shortcomings of EP}
\label{sec:pathologicalresults}

\begin{figure}
\centering
\hspace{0.0cm}
\includegraphics[width=6in]{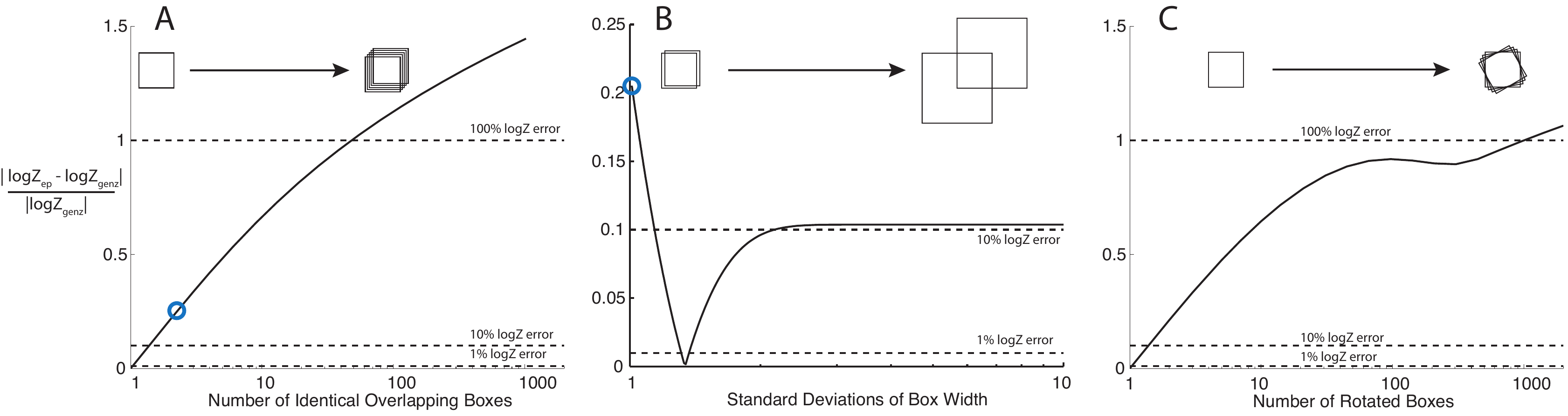}
\caption{\small{Empirical results from intentionally pathological cases that explore the shortcomings of the EP approach.  All panels show the problem of integrating $\mathcal{N}(0,I)$ Gaussian in $n=2$ dimensions over the $[-1,1] \times [-1,1]$ box.  Panel depicts the error as repeated box factors are added to the problem.  The true integral $F(\regionA)$ is unchanged, but the EP error grows arbitrarily bad.  Here EP consistently \emph{underestimates} the true $F(\regionA)$.  Panel B depicts the error as extra factor mass is added.  Though the integration region is unchanged (still the $[-1,1] \times [-1,1]$ box that is the intersection of larger boxes), the factors have extra mass truncated by other factors.  EP can not ignore this extra mass and categorically \emph{overestimates} the true probability.  Here it is important to note that the error in Panel B does not start at 0 (when there is no extra mass), because that case corresponds to 2 repeated boxes.  This case is shown as blue dots in Panels A and B to make that connection (the blue dots are the same problem).  The error starts as an underestimate, and it crosses 0 error at the point at which the redundancy effect (underestimator from Panel A) and extra mass (overestimator from Panel B) compensate for each other.  Both Panels A and B can be ``fixed" by removing redundant or inactive constraints: minimalising the polyhedron as described in Appendix \ref{sec:minpoly}.  However, Panel C shows that these redundancy and extra mass problems can not always be corrected.  Here we show rotated boxes (each rotated by $\pi/1000$ with respect to the previous box), which define the region uniquely and with no inactive constraints.  There are still significant errors due to these effects.  Panel C underestimates the probability, which is sensible because there is more redundancy than there is extra mass in this pathological case.  Though all these examples are contrived, they illustrate serious empirical issues with EP that are not widely discussed.}}
\label{fig:resultspathology} 
\end{figure}

The above results already suggest that certain Gaussian probabilities may be ill-suited to EP.   In particular, we gained intuition from viewing the problem in the whitened space, where the colinearity of the integration region can be seen to imply greater or lesser error when using EP.  Now we want to use that intuition to intentionally create pathological cases for the EP method.  Here we will introduce our terminology for the issues that cause EP to give poor results, which will provide a transition to a more theoretical discussion of the shortcomings of EP in the subsequent Discussion.  Figure \ref{fig:resultspathology} introduces these pathological cases.  In all cases, we want to integrate the $\mathcal{N}(0,I)$ Gaussian in $n=2$ dimensions over the $[-1,1] \times [-1,1]$ box, which is simply the product of two axis-aligned univariate box functions ($t_1(x_1)= \II\{-1 < x_1 < 1\}$ and $t_2(x_2)= \II\{-1 < x_2 < 1\}$).  We note that this choice of Gaussian and integration region is convenient, but that the effects shown here exist in all integration regions that we have tested, to varying extents.  This case is a common and representative case in terms of error magnitudes.  The $y$-axis shows the errors as usual, and the $x$-axis is a feature chosen to create large errors in the EP method.  

First, when looking at a transformed integration region, it seems plausible that some of the error from EP derives from the fact that approximate factors $\tilde{t}_i(\x)$ are not orthogonal, and thus they may double count the mass in some way when forming the EP approximation $q(\x)$.   An extreme, contrived example of this would be adding two repeats of identical factors $t_i(\x)$.  Though the integration region is unchanged, EP is not invariant to this change, as it still must make an approximate factor for each true factor.  This fact has been noted previously in Power-EP \cite[]{PowerEP}, a connection which we explore deeply in the Discussion.   Panel A describes this pathological case, which we call the ``redundancy" issue.   As the cartoon in Panel A depicts, along the $x$-axis we use EP to solve the same probability with increasing numbers of redundant factors.  We begin with a single box at left, where EP gets the exact answer as expected, since this problem decomposes to a product of univariate computations.  Moving right, we add up to 1000 copies of the same integration region, and we see that the error grows to an answer that is arbitrarily bad.  In mathematical terms, we go from using EP to solve $\int p_0(\x) t(x_1)t(x_2) d\x$, to using EP to solve $\int p_0(\x) t(x_1)^{1000}t(x_2)^{1000} d\x$.  Though the true $F(\regionA)$ is unchanged, EP returns a wildly different answer.    More specifically, EP in this redundancy case is \emph{underestimating} the true probability, which is expected and fully explained in the Discussion.  

Second, by returning again to the transformed integration region, it also seems plausible that some of the error from EP derives from the approximate factors accounting for mass inappropriately.  For example, when two $t_i(\x)$ are not orthogonal, the moment matching of the cavity and the factor  to the cavity and the approximate factor will include mass that is outside the integration region.  Panel B of Figure \ref{fig:resultspathology} elucidates this concept.   The integration region of interest here is  still the $[-1,1] \times [-1,1]$ box, but that can be described by the intersection two boxes of any size, as the cartoon in Panel B shows.   However, during the EP iterations, EP must consider the mass that exists in each true box factor \emph{individually}.  Because the Gaussian family has infinite support, there will always be nonzero mass outside the desired intersection that is  the $[-1,1] \times [-1,1]$ box, and EP approximate factors will incorporate this mass.  Hence we expect that EP will not be invariant to this change, and further that EP should overestimate the true probability when there is ``extra mass," which is the term we use to describe this pathology of EP.  Indeed Panel B shows precisely this.  As extra mass is increased along the $x$-axis, error increases and overestimates the true probability.  There are two interesting features here: first, note that the case where there is no extra mass - at left - is still a case where there are two redundant boxes, and hence we expect the underestimation error from Panel A (these corresponding points are circled in blue in Panels A and B).   As extra mass is added, the EP result increases, the error crosses zero, and the EP result becomes a significant overestimate of the true probability.  The second interesting feature is that the error levels off more quickly than in the redundancy case.  This is not surprising, as the exponential decay of the Gaussian implies that the extra mass issue can not arbitrarily punish EP - adding larger and larger boxes gives diminishing additional probability mass that EP must incorporate.

Finally, one might suggest that the problems of Panels A and B are entirely avoidable with appropriate preprocessing.  Specifically, we can pull in all inactive polyhedral constraints, which would remove the extra mass issue of Panel B.  Further, we can prune all redundant constraints, which would remove the redundancy issue of Panel A.   We have discussed this previously in terms of a minimal representation of the polyhedron $\regionA$, and these steps are detailed in Appendix \ref{sec:minpoly}.  However, these trimming and pruning operations will not remove these issues in anything other than these specific cases, as can be seen in Panel C.  Here we essentially replicate the result of Panel A.  However, instead of precisely repeating the box, we add a sequence of slightly rotated boxes.  Thus, this polyhedron is a minimal description of the integration region, and yet still we have identical issues as Panel A and B.  One can imagine a series of heuristic choices to circumvent this issue, but the fact remains that EP will produce poor results for many simpler cases (see Figures \ref{fig:generalresults} and \ref{fig:specialresults}, for example the trapezoid case), and thus we feel an investigation of these error modes is warranted and an interesting EP contribution.

From this figure the reader might correctly suggest that both the redundancy issue of Panel A and the extra mass issue of Panel B are two sides of the same coin: EP enforces a Gaussian approximation to a hard box-function factor, which is a highly non-Gaussian form.  Thus redundancy and extra mass generally go hand in hand: if there are constraints that are not orthogonal, they will each have to consider both mass already considered by another factor (redundancy) and mass that lies outside the actual polyhedron (extra mass).   Nonetheless, keeping these two ideas separate in these results will help us consider the theoretical underpinnings and point towards possible ways to improve the EP estimate of Gaussian probability and EP more generally, which is the subject of the Discussion section.

\subsection{Final note: other moments}

A novelty of our work here is that we have brought an approximate inference framework to bear on the problem of approximate integration.   While the problem of calculating Gaussian probabilities - the zeroth moment of $p(\x)$ - is interesting in its own right, many researchers will ask how well other moments, namely the ``posterior" mean and covariance  of the approximation $q(\x)$ match the true moments.  Throughout our results we calculated these moments with EP also.  While demonstrating this claim rigorously is beyond the scope of this work (which focuses on the zeroth moment), we report anecdotally that we see similar accuracy and error regimes for the mean and covariance estimates from EP.    As Genz methods do not return these moments, we calculated posterior means and covariances with elliptical slice sampling \cite[]{murrayESS}, which further complicates this claim due to the nontrivial error rate and computational burden of this sampling method.   Nonetheless, to answer this common question and broaden the implications of these results beyond simply the zeroth moment, we note our finding that other EP moments follow similar trends.

\section{Discussion}
\label{sec:discussion}

This paper has introduced a number of concepts and results around both Gaussian probabilities and Expectation Propagation.  Here we discuss these findings.

\subsection{EP allows approximate integration and can be used to calculate Gaussian probabilities}

At the highest level, this paper introduces a novel approach to calculating multivariate Gaussian probabilities, which are a difficult and important problem throughout science and applied statistics.   We used an approximate inference framework - Expectation Propagation - in an unexpected way to perform this Gaussian integration.  From the perspective of Gaussian probabilities, this work introduces EPMGP as a method that can be used to give good, often highly accurate, numerical approximations.  Our results show that under a variety of circumstances, relative EP errors rarely exceed $1\%$, and median errors are typically two to four orders of magnitude smaller.  While we spent much of the results exploring the regions of poor performance (which we continue to explore below), the general strength of the method should not be neglected.  The EPMGP algorithm also has some nice features such as analytical derivatives, attractive computational properties, and a natural ability to compute tail probabilities.  On the other hand, existing numerical methods, in particular the Genz method, typically demonstrate high numerical accuracy and a clear runtime/accuracy tradeoff, which we have also demonstrated here.   Furthermore, as we noted, one can readily construct cases wherein the EP method will perform poorly.  Thus we again note that the purpose here is not to compare EPMGP to the Genz method, as those approaches share different properties, but rather to analyse the methods for Gaussian probability calculations across a range of scenarios so that applied researchers can make an informed and practical choice about which methods suit particular needs.   

For machine learning researchers, another value of this work is the use of a popular approximate inference framework for an application other than approximate inference.  Exploring the generality of the machine learning toolset for other purposes is important for understanding broader implications of our field's technologies.

Perhaps the most interesting aspect of this work is that the Gaussian probability problem, by its simple geometric interpretation, allows clear investigation into the strengths and weaknesses of EP.  This work helps us explore empirically and theoretically the circumstances of and reasons behind EP's sometimes excellent and sometimes poor performance, which bears relevance to applications well beyond just Gaussian probabilities.  

\subsection{The strengths and weaknesses of EP: empirical evidence}

Section \ref{sec:results} presents extensive empirical results of the performance of EP for the Gaussian probability problem.  Figures \ref{fig:axisresults} and \ref{fig:generalresults} indicate that the EP algorithm works well for hyperrectangular regions but has considerably lower accuracy (one to two orders of magnitude) for general polyhedral regions.  While this discrepancy may seem fundamental, we showed in Figure \ref{fig:white} that in fact hyperrectangular and polyhedral regions are closely related in a whitened space, where all but trivial cases are non-rectangular polyhedra.  We have introduced the intuitive concepts of extra mass and redundancy, which suggest why EP should be error prone as the integration region (in the whitened space) gets further from rectangular.    Figure \ref{fig:generalresultspoly} provided a first example of this feature.  For a fixed dimensionality $n$, increasing the number of polyhedral faces $m$ increases the redundancy, since each subsequently added constraint will have increasing colinearity with the existing constraint set.  Panel A of Figure \ref{fig:generalresultspoly} shows this increasing error as a function of $m$.  We also investigated cases where the Gaussian problem has special structure, such as orthant and trapezoid integration regions in Figure \ref{fig:specialresults} and equicorrelated Gaussians in Figure \ref{fig:resultsequicorr}.  Focusing now on the redundancy and extra mass issues that we have introduced, Panel B of Figure \ref{fig:specialresults} and all of Figure \ref{fig:resultsequicorr} are particularly useful.  They both demonstrate considerable EP error trend as the Gaussian moves away from white.  Considered from the perspective of the space whitened with respect to the Gaussian, these figures demonstrate that EP error trends reliably with how non-rectangular or coloured the integration region is.  We can concretely see the extra mass and redundancy issues causing errors in this view.  Finally, Figure \ref{fig:resultspathology} presents pointed (albeit contrived) examples to clarify the effect of these issues.   

In total, our empirical evidence suggests that EP can often be used to calculate Gaussian probabilities quickly and to high accuracy, and further that consideration of the Gaussian and integration region at hand can help predict when this computation will not be reliable.  We introduced the intuitive concepts of redundancy and extra mass, which are helpful in understanding why problems come about.  We attempt to condense these intuitions into a more rigorous theoretical understanding in the following section.

\subsection{The strengths and weaknesses of EP: theoretical underpinnings}

We now delve into a more theoretical investigation of EP, first by considering a factor graph representation, and second by considering the divergences that EP is trying to minimise in various problem settings.

\subsubsection{Factorisation perspective}

\begin{figure}
\centering
\hspace{0.0cm}
  \begin{tikzpicture}[node distance=2cm]
    \node[anchor=north west] at (-10,2.5) {\LARGE{\bf A}};
    \node[fac] (p1) at (-8.5,1) {}; 
    \node[anchor=south] at (p1.north) {$\mathcal{N}(x_1) $}; 
    \node[var, below of=p1] (x1) {$x_1$} edge (p1);
   \node[fac, below of=x1] (t1) {} edge (x1);
    \node[anchor=north] at (t1.south) {$t_1(x_1)$};
    \node[fac] (p2) at (-6.5,1) {}; 
    \node[anchor=south] at (p2.north) {$\mathcal{N}(x_2) $}; 
    \node[var, below of=p2] (x2) {$x_2$} edge (p2);
   \node[fac, below of=x2] (t2) {} edge (x2);
    \node[anchor=north] at (t2.south) {$t_2(x_2)$};
	\node[anchor=north west] at (-5.7,-1) {\huge{...}};
    \node[fac] (pn) at (-4,1) {}; 
    \node[anchor=south] at (pn.north) {$\mathcal{N}(x_n) $}; 
    \node[var, below of=pn] (xn) {$x_n$} edge (pn);
   \node[fac, below of=xn] (tn) {} edge (xn);
    \node[anchor=north] at (tn.south) {$t_n(x_n)$};
	%
    \node[anchor=north west] at (-2.5,2.5) {\LARGE{\bf B}};
    \node[fac] (pb1) at (-1,1) {}; 
    \node[anchor=south] at (pb1.north) {$\mathcal{N}(x_1) $}; 
    \node[var, below of=pb1] (xb1) {$x_1$} edge (pb1);
    \node[fac] (pb2) at (1.5,1) {}; 
    \node[anchor=south] at (pb2.north) {$\mathcal{N}(x_2) $}; 
    \node[var, below of=pb2] (xb2) {$x_2$} edge (pb2);
    \node[fac] (pb3) at (4,1) {}; 
    \node[anchor=south] at (pb3.north) {$\mathcal{N}(x_3) $}; 
    \node[var, below of=pb3] (xb3) {$x_3$} edge (pb3);
	   \node[fac] (tb1) at (0.25, 0) {} edge (xb1) edge (xb2);
	    \node[anchor=south] at (tb1.north) {$t_1(x_1,x_2)$};
	   \node[fac] (tb2) at (2.75, 0) {} edge (xb2) edge (xb3);
	    \node[anchor=south] at (tb2.north) {$t_2(x_2,x_3)$};
	   \node[fac, below of=xb3] (tb3) {} edge (xb3);
    	  \node[anchor=north] at (tb3.south) {$t_3(x_3)$};
	   \node[fac] (tb4) at (1.0, -3) {} edge (xb1) edge (xb2) edge (xb3);
	    \node[anchor=north] at (tb4.south) {$t_4(x_1,x_2,x_3)$};
  \end{tikzpicture}
\caption{\small{Factorisation perspective. Panel A represents the factor graph for an axis-aligned region in the whitened space ($\mathcal{N}(x_i)$ is a standard Gaussian on $x_i$, which defines the whitened prior $p_0'(\x)$). Panel B represents an example of a region which is not axis-aligned in the whitened space: we show a polyhedral region with $n=3$ which had a bidiagonal prior covariance structure, yielding dependent factors in the whitened space.}}
\label{fig:factorization-view} 
\end{figure}
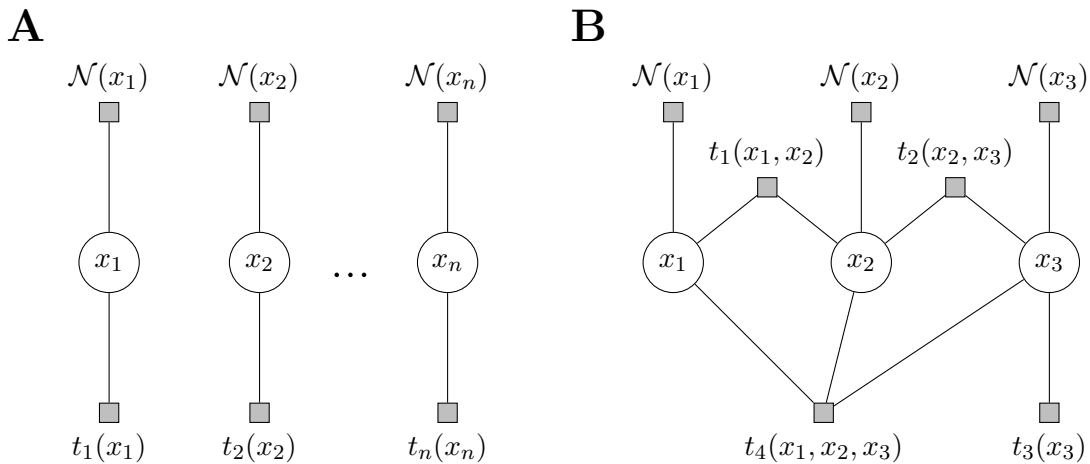

As we explained in section~\ref{sec:white}, we can equivalently represent any hyperrectangular or polyhedral integration region as a polyhedral region in the whitened space where the Gaussian has identity covariance. This standardisation enables us to focus on the geometry of the region.  If this transformed region is actually axis-aligned, we obtain a fully factorised distribution. The factor graph representing this situation is shown in Panel A of Figure~\ref{fig:factorization-view}, where we use one node for each dimension to explicitly show the factorisation. Because the distribution is fully factorised, the integration also factorises as a product of univariate integrals, and so local moment matching in this case also yields global moment matching.   Mathematically, the global KL-divergence can now be directly related to the local divergences such that $D_\KL(p\| q) = \sum_i D_\KL(t_i p'_{0i}\| \tilde{t}_i p'_{0i})$, where $p'_{0i}$ is the $i$th dimension of the whitened Gaussian $p'_0(\x)$ (standard univariate normals as in Figure \ref{fig:factorization-view}).  Accordingly, the KL minimisation problem separates, and EP will yield the exact zeroth moment (to within the machine precision of the error function).  To be clear, EP is still making an approximation (the Gaussian approximation $q$), but we know that this $q$ will have the exact zeroth, first, and second moments of $p$.  And thus the Gaussian probability will be correct.

On the other hand, non axis-aligned constraints in the whitened space can be seen as linking multiple nodes together, thereby inducing dependencies in the factor graph and breaking the equivalence between local moment matching and global moment matching. These non axis-aligned constraints could either arise from the original integration region where a constraint links variables together, or from a non-diagonal Gaussian covariance which colours the integration region when space is whitened with respect to that covariance.  We show an example of these dependencies in Panel B of Figure~\ref{fig:factorization-view}. This factor graph is obtained by taking a bidiagonal prior covariance matrix on a polyhedral region for $n=3$ and transforming it to the whitened space as in Figure \ref{fig:white}. The polyhedral region is an axis-aligned hyperrectangle plus an additional factor constraint that links all dimensions.  The covariance matrix induces pairwise factors in the whitened space, and the polyhedral constraint induces a factor on all variables. These non axis-aligned factors give rise to loops\footnote{We note in passing that the EP factor formulation that we are using is different than the standard one used on factor graphs which replaces each factor with a product of univariate approximate factors -- see for example Section 10.7.2 of~\cite{bishopBook} -- and which reduces to loopy belief propagation in the case of discrete variables. Our formulation keeps the full approximate factor $\tilde{t}_i(\x)$, much as when EP is applied on the Bayesian posterior of an i.i.d. model.} in the factor graph which we expect to reduce the accuracy of EP, as seen in Figure~\ref{fig:generalresults} for polyhedral regions.

Little is formally known about the accuracy of EP on loopy graphs.  In the case of Gaussian probabilities, fortunately we can use the geometry of the integration region to interpret the effect of additional graph edges. In particular, factors having several variables in common correspond to constraints which have a non-zero dot product (somewhat co-linear).  Thus, these factors create approximate redundancy (Panel~C of Figure~\ref{fig:resultspathology}) and extra mass (Panel~B of the same figure).   While these EP issues exist in all EP problem setups (not only Gaussian probabilities), the geometrical concepts available here provide a unique opportunity to understand the accuracy of EP which would not be apparent in standard factor graphs.  We formalise these concepts in the next section.  

\subsubsection{Alpha-divergence perspective}

The redundancy and extra mass problems can be viewed in a unifying way as changing the effective \emph{alpha-divergence} that EP is minimising, as we explain in this section. The alpha-divergence $D_{\alpha}(p \parallel q)$ is a generalisation of the KL-divergence.  The relationship of alpha-divergences to EP and other variational schemes has been studied: \cite{minkaMSFTTR2005} shows that several message passing algorithms on factor graphs can be interpreted as the local minimisation of different alpha-divergences. Following the notation from~\cite{minkaMSFTTR2005}, $\KL (q \parallel p)$, the Hellinger distance, and $\KL (p \parallel q)$ corresponds to $\alpha=0$, $\alpha=0.5$, and $\alpha=1$ respectively. Hence EP can be interpreted as the minimisation of local alpha-divergences with $\alpha=1$. 

The Power-EP algorithm~\cite[]{PowerEP}\footnote{We warn the reader that the notation for alpha-divergence varies between papers (and in particular between~\cite{PowerEP} and~\cite{minkaMSFTTR2005}). We use the notation from~\cite{minkaMSFTTR2005} -- the corresponding $\alpha'$ used in~\cite{PowerEP} is obtained as $\alpha' = 2 \alpha - 1$.} is an extension of EP where the factors are raised to a specific power $\alpha_i$ before doing moment matching. One of the motivations for this algorithm is that raising the factor $t_i$ to a power can make the projection more tractable.  That paper also notes that running EP on a model where each factor $t_i$ is repeated $n_i$ times is equivalent to running Power-EP (with the factors unrepeated) with $\alpha_i = 1/n_i$.  Importantly, the Power-EP algorithm can also be derived as the minimisation of local $\alpha_i$-divergences. 

This repetition of factors is precisely the ``redundancy" issue that we created by repeating constraints $n_i$ times (as in Panel~A of Figure~\ref{fig:resultspathology}). Our EP algorithm in the pure redundant case can thus be interpreted as running Power-EP with $\alpha_i = 1/n_i$.  This relationship is useful to us for four reasons. 

First, the zeroth moment $Z$ is always \emph{underestimated} when doing global minimisation of \emph{exclusive} alpha-divergence with $\alpha<1$ \cite[]{minkaMSFTTR2005}.  Also, $Z$ is only correctly estimated for $\alpha=1$ (KL-divergence), and it is \emph{overestimated} for \emph{inclusive} alpha-divergences which have $\alpha>1$.   This provides a theoretical explanation for the systematic underestimation of $Z$ in the redundancy case of Panel~A, Figure~\ref{fig:resultspathology}:  EP in this case is the same as Power-EP with $\alpha_i = 1/n_i < 1$.  As in regular EP, a subtle aspect here is that Power-EP is doing \emph{local} alpha-divergence minimisation, whereas the overestimation/underestimation of the zeroth moment $Z$ holds for \emph{global} alpha-divergence minimisation, and the relationship between the two is not yet fully understood.   However, our construction in Panel A of Figure \ref{fig:resultspathology} was a fully factorised case, so there is direct correspondence between global and local divergences, and as such we did systematically observe underestimation in these experiments.  We also add that our experiments with Panel C of this figure and other similar cases indicate similar underestimation even when there is not direct global-local correspondence. 

Second, this relationship between EP and alpha-divergence gives us a notion of \emph{effective} alpha-divergence $\alpha_\text{eff}$ for each factor. In particular, running EP with the factor $t_i$ repeated $n_i$ times is the same as the local alpha-divergence minimisation by Power-EP with $\alpha_\text{eff} = 1/n_i$. When the constraints are almost-repeated (they have a large dot product, but are not fully colinear, such as in Panel~C of Figure~\ref{fig:resultspathology}), we lose the exact correspondence between EP and Power-EP, but we could still imagine that EP would correspond to Power-EP with a suitable $\frac{1}{n_i} < \alpha_\text{eff}<1$ in this case as well (using a continuity argument). 

Third, we can now think of \emph{correcting} for this $\alpha_\text{eff}$ by using Power-EP with $\alpha' = 1/\alpha_\text{eff}$ instead of standard EP. In the case of exactly repeated constraints, this means using $\alpha'_i = n_i$.  Indeed, our additional experiments (not shown) demonstrate that Power-EP does yield back the correct answer in this simple case. In other words, using this notion of correcting for $\alpha_{\text{eff}}$, the errors of Panel A in Figure \ref{fig:resultspathology} disappear.   In the case of almost-repeated constraints, we can still imagine correcting the effect by choosing a suitable $\alpha_\text{eff}$ correction term and using Power-EP.  Our early investigation of this correction has given some promising results.   An important and interesting direction of future work is characterising $\alpha_\text{eff}$ from geometric quantities such as dot products of constraints. 

Fourth and finally, the extra mass issue can also be viewed in terms of alpha-divergences, though the connection is not as rigorous as the redundancy issue.  The extra mass problem is one of inclusivity: the Gaussian approximate factor is including mass that it should not (as that mass lies outside the true polyhedron), which is a property of \emph{inclusive} alpha-divergences as mentioned previously, {\it i.e.}, divergences with $\alpha > 1$.  These we would expect to overestimate the true probability, and indeed this is what we see in Panel B of Figure \ref{fig:resultspathology}.  We can also consider correcting for these $\alpha_{\text{eff}}$ with Power EP.  The situation is slightly more complicated here as the extra mass problem involves understanding how $\alpha_{\text{eff}}$ changes with the decay of the Gaussian distribution, but our initial attempts at these corrections indicate that the overestimation error can be significantly reduced.

To summarise, both the redundancy and extra mass problems can be cast in the language of effective alpha-divergence. We know that minimising $D_{\alpha}$ only returns moment matching when $\alpha = 1$ (KL-divergence), and thus any efforts to drive $\alpha_{\text{eff}}$ closer to 1 should improve all the performance results seen here.  We have noted our success in removing errors entirely in the simplest cases such as Panel A of Figure~\ref{fig:resultspathology} and in removing errors significantly in other slightly more complicated scenarios.  Thus we feel that a judicious choice of $\alpha$, to correct for the effective $\alpha$ in a Power-EP framework, is a fruitful direction of future work for both the theoretical understanding of and the empirical performance of EP across many applications.

\subsection{Broader implications for EP}


Having delved deeply into EP for Gaussian probabilities, it is worth stepping back to consider the breadth of these findings.

First we draw connections to other common machine learning applications.  In the noise-free Bayes Point Machine model \cite[]{herbrichBook, minkaUAI01}, the model evidence is the integral of a white Gaussian over a polyhedral cone (intersection of halfspaces going through the origin), and so is a special case of our Gaussian probability framework.  EP was analysed for the Bayes Point Machine in \cite{minka01phd,minkaUAI01} and was found to have good performance.  On the other hand, its accuracy was only evaluated on a synthetic example with 3 points in 2 dimensions, which in our terminology would be $\{m=3 , n=2\}$, which forms but a small part of our empirical analyses.   Furthermore, prediction accuracy on a test set was used to evaluate EP on real world examples, but as ground truth was unavailable, the estimation accuracy remains unclear.  \cite{minka01phd} observed our redundancy problem in a synthetic example: the author notes repetitions of the same datapoints (and so repeating the halfspace constraint in our formulation) yielded much poorer approximations of the evidence or the mean.   By comparing to numerical integration methods, we have provided a more extensive analysis of EP's accuracy for this problem.  Further, we anticipate that a method for estimating $\alpha_{\text{eff}}$ (and using Power-EP) could improve the performance of Bayes Point Machines when the data is highly redundant.

Another important existing connection point is Gaussian Process classification \cite[]{rasmussenBook, KussRasmussen2005}.  When a probit likelihood (namely a univariate Gaussian cdf) is used, the data likelihood becomes a Gaussian probability problem.   The same is true of probit regression \cite[]{ochi1984}, where changing the representation of the more standard prior and likelihood reveals a high-dimensional Gaussian probability.  Though these problems are usually not considered as Gaussian probabilities, it is another point of broad importance for EP and for Gaussian probabilities.  Furthermore, as EP is often used in a different way in these problems, exploring the similarities between EP for Gaussian probabilities and EP for problems that can be cast as Gaussian probabilities is an interesting subject for future investigation.

We have revealed some interesting empirical and theoretical properties for EP as applied to approximate Gaussian integration, but understanding how much of these issues stems from EP itself and how much from the application is an important clarification.   Gaussian probabilities are a particularly instructive application, as they encourage the geometric intepretation that we have maintained throughout this paper.  However, the issues of redundancy and extra mass are not specific to strict box functions, and the interpretation in terms of $\alpha_{\text{eff}}$ is no less applicable in the case of smooth functions.   To test this question, we could repeat the setup of Figure \ref{fig:resultspathology} but using smooth functions such as probit curves.  This problem would then bear similarity to Gaussian Process classification, and we could stress test the sensitivity of EP (in terms of extra mass and redundancy) to changing amounts of smoothness in the probit, where one end of this continuum is our Gaussian probability results.  Though this remains a subject of interesting further work, we anticipate a result consistent with our findings here.

\subsection{Future Directions}

This work has explored both Gaussian probabilities and Expectation Propagation, and there are opportunities for valuable future work in both.  We feel the most interesting direction for further investigation regards EP, specifically the theoretical discussion of alpha-divergences as in the section above.  Any insights into understanding $\alpha_{\text{eff}}$ based on the geometry of the distributions involved, and further methods by which to correct for that $\alpha_{\text{eff}}$ via Power-EP approaches, could have broad implications for the continued success of EP as a powerful approximate inference framework.  Here we have explored the geometrically and functionally simple case of box functions (which give Gaussian probablities), but future work should explore this for a variety of other factors (likelihood terms) as well.

The other focus for future work could be improvements to EPMGP specifically.   For example, EP can often be improved by updating multiple approximate factors together.  Here we considered a rank-one factor update, but also highly redundant factors could be considered in a rank-two update together.  Because some of these low dimensional Gaussian probabilities can be calculated (such as orthants), these updates will be tractable in certain cases. 

Finally, it is worth noting that our EP approach to approximate integration highlights the connections between approximate inference and integration, so some interesting connections here could be explored.  For example, we noted that the Genz methods do not currently offer the ability to compute tail probabilities, calculate higher moments than the zeroth, or offer analytical derivatives.   The absence of these features should not in principle be fundamental to the numerical integration approach, so reconsidering the Genz methods in the light of approximate inference could yield novel techniques for machine learning.  Furthermore, hybrid approaches could be considered, drawing strengths from EP and numerical integration approaches.

\section{Conclusion}
\label{sec:conclusion}

We have presented an algorithm constructing analytic approximations to polyhedral integrals on multivariate Gaussian distributions. Such approximations are of value for a number of applications in many fields, particularly in hyperrectangular integration cases. We presented extensive numerical results for this case and the more general polyhedral case.  We showed an interesting result that a small generalisation of the same algorithm is considerably less reliable. We explored the empirical and theoretical reasons behind these performance differences, which elucidated some features of EP not often discussed.  These results also reveal Gaussian Probabilities as a useful testbed for analysing EP.  As Expectation Propagation is becoming a more popular approximate inference method, this cautionary result suggests that EP should not be trusted blindly in new applications. Nonetheless, the uses of this EP framework will hopefully enable future work that uses the important object that is Gaussian probability.

\acks{We thank Tom Minka, Thore Graepel, Ralf Herbrich, Carl Rasmussen, Zoubin
  Ghahramani, Ed Snelson, and David Knowles for helpful discussions. JPC was supported by the UK
  Engineering and Physical Sciences Research Council (EPSRC
  EP/H019472/1); PH and SLJ were supported by a grant from Microsoft Research
  Ltd.}


\newpage

\appendix

\newpage

\appendix

\section{Gaussian EP cavity distributions for rank one factors $t_i(\x)$}
\label{sec:rank1cavity}

The purpose of this appendix is to derive the forms of the EP cavity distributions given in Equation \ref{eqn:cavityepmgp}.  The only specific requirement is that the factors are rank one, that is: $t_i(\x) = t_i(\c_i^T\x)$. We begin with the high dimensional Gaussian cavity distribution:
\begin{equation}
q^{\wo i}(\x) ~~=~~ Z^{\wo i}\mathcal{N}(\x; \u^{\backslash i}, V^{\backslash i}) ~~~\propto~~~ p_0(\x) \prod_{j\ne i} \tilde{t}_j(\x)
\end{equation}
where
\begin{equation}
V^{\backslash i} = \left(K^{-1}  + \sum_{j\ne i} \frac{1}{\tilde{\sigma}_j^2} \c_j\c_j^T\right)^{-1}
\end{equation}
and
\begin{equation}
\u^{\backslash i} = V^{\backslash i}\left(K^{-1}\m  + \sum_{j\ne i} \frac{\tilde{\mu}_j}{\tilde{\sigma}_j^2} \c_j\right)
\end{equation}
using standard properties of the normal distribution ({\it e.g.}, \cite{rasmussenBook}).  Alternatively, as is often done for EP, we can consider subtracting the $i$th approximate factor from the overall posterior approximation $q(\x)$.  That is, since
\begin{equation}
q(\x) = Z\mathcal{N}(\x; \boldmu, \Sigma) ~~~\propto~~~ p_0(\x) \overset{m}{\prod_{j=1}} \tilde{t}_j(\x),
\end{equation}
we can say that $q^{\wo i}(\x)$ has parameters:
\begin{equation}
\label{eqn:Vslash}
V^{\backslash i} = \left(\Sigma^{-1}  - \frac{1}{\tilde{\sigma}_i^2} \c_i\c_i^T\right)^{-1}
\end{equation}
and
\begin{equation}
\label{eqn:mslash}
\u^{\backslash i} = V^{\backslash i}\left(\Sigma^{-1}\boldmu  - \frac{\tilde{\mu}_i}{\tilde{\sigma}_i^2} \c_i\right).
\end{equation}

Note that these steps are just like standard EP, save that we are subtracting rank one factors $\tilde{t}_i(\x)$ instead of the more common axis-aligned factors.  We will use this cavity to moment match $q^{\wo i}(\x)\tilde{t}_i(\x)$ to $q^{\wo i}(\x)t_i(\x)$.  Since our $t_i(\x)$ are rank one, we can equivalently marginalise out all other dimensions and moment match in a single dimension, which simplifies computation and presentation.  This equivalence of marginal moment matching with the high-dimensional moment matching for EP requires that the exponential family used for the approximate factors is closed under marginalisation (which is the case for Gaussians), as noted in Section 3 of \cite{seeger08epexpfam}.  Accordingly, we want to form a univariate cavity distribution by marginalising over all dimensions orthogonal to the univariate factor of interest:
\begin{equation}
q_{\wo i}(\c_i^T\x) = \int_{\backslash \c_i;\x} q^{\wo i}(\x') d\x'
\end{equation}
where the integration region $\backslash \c_i;\x$ indicates marginalisation over all dimensions orthogonal to $\c_i$ (an $n-1$ dimensional integral) and we move the cavity-defining ``$\wo i$" superscript down to the subscript to indicate going from high dimension to single dimension.  In standard EP, this notation would be $\backslash \e_i$ for the axis defined by the unit vector $\e_i$; in other words, we marginalise all variables but the $i$th.  Mathematically, $\backslash \c_i;\x$ is the $n-1$ dimensional affine space perpendicular to $\c_i$ and defined by a particular choice of $\c_i^T\x$, that is $\backslash \c_i;\x = U_{\c_i} + (\c_i^T\x) \c_i$ and $U_{\c_i} = \bigl\{ \x'\in\reals^n \mid \c_i^T\x' = 0 \bigr\}$ (note that $\c_i$ is assumed to be normalised).

To do this marginalisation in closed form, we do a simple change of variable. We exploit the fact that, for any unit norm vector $\c_i$, there exists an orthonormal basis $\{\c_i, \a_2,...,\a_n\}$ of $\reals^n$.  We define the orthogonal matrix
\begin{equation}
A = \begin{bmatrix} \c_i & \a_2 & \hdots & \a_n  \end{bmatrix}, ~~~~\mathrm{with}~~~~  A^TA = AA^T = I,
\end{equation}
and change the coordinates for our integration of $q^{\wo i}(\x)$ with the change of variable $\y = A^T \x'$. Since $A$ is orthogonal, the Jacobian determinant is one. The Gaussian $q^{\wo i}(\x')= Z^{\wo i} \mathcal{N}(\x'; \u^{\wo i},V^{\wo i})$ simply becomes $Z^{\wo i} \mathcal{N}(\y; A^T \u^{\wo i},A^T V^{\wo i} A)$, and so:
\begin{eqnarray}
q_{\wo i}(\c_i^T\x) & = & \int_{\backslash \c_i;\x} q^{\wo i}(\x') d\x' \\
& = & \int_{\backslash \e_1;(\c^T\x)\e_1}  Z^{\wo i} \mathcal{N}(\y; A^T \u^{\wo i},A^T V^{\wo i} A) d\y \\
& = & Z^{\wo i} \mathcal{N}(\c_i^T \x; \c_i^T\u^{\backslash i} , \c_i^TV^{\backslash i}\c_i)
\end{eqnarray}
where the second to last equality follows from the fact that integrating out all dimensions orthogonal to $\c_i$ is equivalent to marginalising all but the first axis ($\e_1$ of $A^T\x$), by our definition of the matrix $A$.  Since those orthogonal dimensions all normalise to 1, the high dimensional integral becomes univariate.  The last equality follows again from the definition of $A$, and thus we are left with a simple univariate cavity distribution.  By using the matrix inversion lemma and the definitions from Equations \ref{eqn:Vslash} and \ref{eqn:mslash}, we can derive simple  forms for these cavity parameters:

\begin{eqnarray}
\sigma_{\wo i}^{2} & = &  \c_i^TV^{\backslash i}\c_i \\
& = & \c_i^T(\Sigma^{-1}  - \frac{1}{\tilde{\sigma}_i^2} \c_i\c_i^T)^{-1}\c_i \\
& = & \c_i^T(\Sigma + \Sigma \c_i ( \tilde{\sigma}_i^{2}  - \c_i^T \Sigma \c_i )^{-1} \c_i^T \Sigma) \c_i \\
& = & \c_i^T\Sigma\c_i\Bigl( 1 +  \frac{ \c_i^T \Sigma \c_i }{ \tilde{\sigma}_i^{2}  - \c_i^T \Sigma \c_i }\Bigr)\\
& = & \frac{\tilde{\sigma}_i^{2}\c_i^T\Sigma\c_i}{\tilde{\sigma}_i^{2}  - \c_i^T \Sigma \c_i }\\
& = & \bigl( (\c_i^T\Sigma \c_i)^{-1} - \tilde{\sigma}^{-2}_i\bigr)^{-1}
\end{eqnarray}

\noindent which matches the form of Equation \ref{eqn:cavityepmgp}.  So too, for the mean parameter of the univariate cavity distribution, we have:

\begin{eqnarray}
\mu_{\wo i} & = & \c_i^T\u^{\backslash i} \\
& = & \c_i^TV^{\backslash i}(\Sigma^{-1}\boldmu  - \frac{\tilde{\mu}_i}{\tilde{\sigma}_i^2} \c_i) \\
& = & \c_i^TV^{\backslash i}\Sigma^{-1}\boldmu - \frac{\tilde{\mu}_i}{\tilde{\sigma}_i^2} \sigma_{\wo i}^2 \\
& = & \c_i^T(\Sigma + \Sigma \c_i ( \tilde{\sigma}_i^{2}  - \c_i^T \Sigma \c_i )^{-1} \c_i^T \Sigma) \Sigma^{-1}\boldmu  - \frac{\tilde{\mu}_i}{\tilde{\sigma}_i^2} \sigma_{\wo i}^2  \\
& = & \c_i^T\boldmu + \c_i^T\Sigma \c_i ( \tilde{\sigma}_i^{2}  - \c_i^T \Sigma \c_i )^{-1} \c_i^T\boldmu - \frac{\tilde{\mu}_i}{\tilde{\sigma}_i^2} \sigma_{\wo i}^2 \\
& = & \c_i^T\boldmu\Bigl( 1 +  \frac{ \c_i^T \Sigma \c_i }{ \tilde{\sigma}_i^{2}  - \c_i^T \Sigma \c_i }\Bigr) - \frac{\tilde{\mu}_i}{\tilde{\sigma}_i^2} \sigma_{\wo i}^2\\
& = & \frac{\c_i^T\boldmu}{\c_i^T\Sigma\c_i} \sigma_{\wo i}^2 - \frac{\tilde{\mu}_i}{\tilde{\sigma}_i^2} \sigma_{\wo i}^2\\
& = & \sigma_{\wo i}^2 \Bigl( \frac{\c_i^T\boldmu}{\c_i^T\Sigma\c_i} - \frac{\tilde{\mu}_i}{\tilde{\sigma}_i^2} \Bigr),
\end{eqnarray}

\noindent which again matches the result of Equation \ref{eqn:cavityepmgp}.  These cavity parameters $\{\mu_{\wo i},\sigma_{\wo i}^2\}$ are intuitively sensible.  Comparing to the standard EP cavity form ({\it e.g.}, \cite{rasmussenBook}), we see that instead of using the axis aligned univariate marginal defined by $\{\mu_i,\sigma^2_i\}$ (the $i$th entry and diagonal entry of $\mu$ and $\Sigma$, the parameters of the approximation $q(\x)$), the cavities over arbitrary polyhedra require the univariate projection onto the axis $\c_i$.

\newpage

\section{Efficient and stable EPMGP implementation}
\label{sec:epmgpstable}

EP gives us a simple form for calculating the parameters of the approximation $q(\x)$ (from which we use the normalising constant $Z$ as the Gaussian probability approximation).   However, there are two potential problems that deserve investigation.  First, each update of $\{\boldmu,\Sigma\}$ naively requires $\mathcal{O}(n^3)$ computation, which may be needlessly burdensome in terms of run time and may introduce numerical instabilities.  Second, as factor parameters $\tilde{\sigma}_i^2$ can tend to infinity ({\it i.e.}, that factor $t_i(\x)$ has no effect on the MGP calculation), we want to reparametrise the calculation for numerical stability.   This reparameterisation is also critical for calculating tail probabilities.

We introduce the natural parameters $\tilde{\tau}_i = \frac{1}{\tilde{\sigma}_i^2}$ and $\tilde{\nu}_i = \frac{\tilde{\mu}_i}{\tilde{\sigma}_i^2}$, and we rewrite Equation \ref{eqn:cavityepmgp} in terms of these parameters, as is standard for EP.  In particular the cavity parameters are calculated as:
\begin{eqnarray}
\label{eqn:cavityepmgpnatural}
\mu_{\wo i} & = &  \sigma_{\wo i}^2 \Bigl( \frac{\c_i^T\boldmu}{\c_i^T\Sigma\c_i} - \tilde{\nu}_i \Bigr)
~~~\mathrm{and}~~~
\sigma^2_{\wo i} = \bigl( (\c_i^T\Sigma \c_i)^{-1} - \tilde{\tau}_i\bigr)^{-1}.
\end{eqnarray}
We note that we do not reparameterise the cavity parameters (just the site parameters within that calculation), as the cavity parameters remain well behaved and numerically stable.  Next, we rewrite the parameters of $q(\x)$ from Equation \ref{eqn:epmgppost} as  
\begin{equation}
\boldmu  = 
\Sigma\Bigl(K^{-1}\m + \overset{m}{\sum_{i=1}}\tilde{\nu}_i\c_i \Bigr)~~~~~~\mathrm{and}~~~~~\Sigma = \Bigl(K^{-1} +
\overset{m}{\sum_{i=1}} \tilde{\tau}_i\c_i\c_i^T\Bigr)^{-1}
\end{equation}

Again these updates have cubic run time complexity.  Let $\Delta\tilde{\tau}_i = \tilde{\tau}_i^{\mathrm{new}} - \tilde{\tau}_i^{\mathrm{old}}$.  Now we consider updating $\Sigma_{\mathrm{old}}$ to $\Sigma_{\mathrm{new}}$ with an update  $\Delta\tilde{\tau}_i$:

\begin{eqnarray}
\label{eqn:sigmanew}
\Sigma_{\mathrm{new}} & = & \Bigl(K^{-1} + \overset{m}{\sum_{j\ne i}} \tilde{\tau}_j^{\mathrm{old}}\c_j\c_j^T + \tilde{\tau}_i^{\mathrm{new}}\c_i\c_i^T \Bigr)^{-1} \\
& = & \Bigl(K^{-1} + \overset{m}{\sum_{i=1}} \tilde{\tau}_i^{\mathrm{old}}\c_i\c_i^T + \Delta\tilde{\tau}_i\c_i\c_i^T \Bigr)^{-1} \\
& = & (\Sigma_{\mathrm{old}}^{-1} + \Delta\tilde{\tau}_i\c_i\c_i^T )^{-1} \\
& = & \Sigma_{\mathrm{old}} - \Sigma_{\mathrm{old}} \c_i \Bigl(\frac{1}{\Delta\tilde{\tau}_i} + \c_i^T \Sigma_{\mathrm{old}} \c_i\Bigr)^{-1} \c_i^T\Sigma_{\mathrm{old}} \\
& = & \Sigma_{\mathrm{old}} - \Biggl(\frac{\Delta\tilde{\tau}_i }{ 1 +  \Delta\tilde{\tau}_i \c_i^T\Sigma_{\mathrm{old}}\c_i }\Biggr) (\Sigma_{\mathrm{old}} \c_i)(\Sigma_{\mathrm{old}} \c_i)^T,
\end{eqnarray}

\noindent which is conveniently just a rank one (quadratic run time) update of the previous covariance.  By a similar set of steps, we derive an update the mean of $q(\x)$, which gives:

\begin{equation}
\label{eqn:munew}
\boldmu_{\mathrm{new}} = \boldmu_{\mathrm{old}} +    \Biggl( \frac{\Delta\tilde{\nu}_i - \Delta\tilde{\tau}_i \c_i^T\boldmu_{\mathrm{old}} }{ 1 + \Delta\tilde{\tau}_i \c_i^T\Sigma_{\mathrm{old}} \c_i} \Biggr) \Sigma_{\mathrm{old}} \c_i,
\end{equation}

\noindent which again involves only quadratic computations and uses numerically stable natural parameters for the approximate factor $\tilde{t}_i(\x)$ terms.

The final term that warrants consideration is the normalisation constant $Z$, which in fact is our approximation to the high dimensional Gaussian probability and the entire point of the EPMGP algorithm.  Starting from Equation \ref{eqn:logZ} and including the definition of $\tilde{Z}_i$ from Equation \ref{eqn:site}, that term is:

\begin{eqnarray}
\log Z &= & 
-\frac{1}{2}\log\lvert K \rvert 
-\frac{1}{2}\overset{m}{\sum_{i=1}} \log \tilde{\sigma}^2_i
+ \frac{1}{2}\log\lvert \Sigma \rvert 
-\frac{1}{2}\m^TK^{-1}\m \nonumber 
+\frac{1}{2}\boldmu^T\Sigma^{-1}\boldmu
- \frac{1}{2}\overset{m}{\sum_{i=1}}\frac{\tilde{\mu}_i^2}{\tilde{\sigma}_i^2}  
\nonumber \\ && ~~
+ \overset{m}{\sum_{i=1}}\log
\tilde{Z}_i - \frac{m}{2}\log(2\pi) \nonumber \\
& = & 
-\frac{1}{2}\log\lvert K \rvert 
-\frac{1}{2}\overset{m}{\sum_{i=1}} \log \tilde{\sigma}^2_i
+ \frac{1}{2}\log\lvert \Sigma \rvert 
-\frac{1}{2}\m^TK^{-1}\m \nonumber 
+\frac{1}{2}\boldmu^T\Sigma^{-1}\boldmu
- \frac{1}{2}\overset{m}{\sum_{i=1}}\frac{\tilde{\mu}_i^2}{\tilde{\sigma}_i^2}\nonumber
\\ && ~~
+ \overset{m}{\sum_{i=1}}\log
\hat{Z}_i + \frac{m}{2}\log(2\pi) + \frac{1}{2}\overset{m}{\sum_{i=1}}\log (\sigma_{\wo i}^2 + \tilde{\sigma}^2_i) 
+ \frac{1}{2}\overset{m}{\sum_{i=1}}\frac{(\mu_{\wo i} - \tilde{\mu}_i)^2}{(\sigma_{\wo i}^2 + \tilde{\sigma}^2_i)}- \frac{m}{2}\log(2\pi) \nonumber \\
& = & 
-\frac{1}{2}\log\lvert K \rvert 
-\frac{1}{2}\overset{m}{\sum_{i=1}} \log \tilde{\sigma}^2_i
+ \frac{1}{2}\log\lvert \Sigma \rvert 
-\frac{1}{2}\m^TK^{-1}\m \nonumber 
+\frac{1}{2}\boldmu^T\Sigma^{-1}\boldmu
- \frac{1}{2}\overset{m}{\sum_{i=1}}\frac{\tilde{\mu}_i^2}{\tilde{\sigma}_i^2}\nonumber
\\ && ~~
+ \overset{m}{\sum_{i=1}}\log
\hat{Z}_i + \frac{1}{2}\overset{m}{\sum_{i=1}}\log (\sigma_{\wo i}^2 + \tilde{\sigma}^2_i) 
+ \frac{1}{2}\overset{m}{\sum_{i=1}}\frac{(\mu_{\wo i} - \tilde{\mu}_i)^2}{(\sigma_{\wo i}^2 + \tilde{\sigma}^2_i)} \nonumber \\
& = & 
-\frac{1}{2}\log\lvert K \rvert 
+ \frac{1}{2}\log\lvert \Sigma \rvert 
-\frac{1}{2}\m^TK^{-1}\m \nonumber 
+\frac{1}{2}\boldmu^T\Sigma^{-1}\boldmu
- \frac{1}{2}\overset{m}{\sum_{i=1}}\frac{\tilde{\mu}_i^2}{\tilde{\sigma}_i^2}
\nonumber \\ && ~~
+ \overset{m}{\sum_{i=1}}\log
\hat{Z}_i + \frac{1}{2}\overset{m}{\sum_{i=1}}\log (1 + \tilde{\sigma}^{-2}_i\sigma_{\wo i}^2) 
+ \frac{1}{2}\overset{m}{\sum_{i=1}}\frac{(\mu_{\wo i} - \tilde{\mu}_i)^2}{(\sigma_{\wo i}^2 + \tilde{\sigma}^2_i)} \nonumber \\
& = & 
-\frac{1}{2}\log\lvert K \rvert 
+ \frac{1}{2}\log\lvert \Sigma \rvert 
-\frac{1}{2}\m^TK^{-1}\m 
+\frac{1}{2}\boldmu^T\Sigma^{-1}\boldmu
\nonumber \\ && ~~
+ \overset{m}{\sum_{i=1}}\log
\hat{Z}_i + \frac{1}{2}\overset{m}{\sum_{i=1}}\log (1 + \tilde{\sigma}^{-2}_i\sigma_{\wo i}^2) 
+ \frac{1}{2}\overset{m}{\sum_{i=1}}\frac{(\mu_{-i}^2 - 2\mu_{\wo i}\tilde{\mu}_i - \tilde{\mu}^2_i\tilde{\sigma}^{-2}_i\sigma^2_{\wo i})\tilde{\sigma}^{-2}_i}{(1 + \tilde{\sigma}^{-2}_i\sigma_{\wo i}^2) } 
\nonumber \\
& = & 
-\frac{1}{2}\log\lvert K \rvert 
+ \frac{1}{2}\log\lvert \Sigma \rvert 
-\frac{1}{2}\m^TK^{-1}\m \nonumber 
+\frac{1}{2}\boldmu^T\Sigma^{-1}\boldmu
\\ && ~~
+ \overset{m}{\sum_{i=1}}\log
\hat{Z}_i + \frac{1}{2}\overset{m}{\sum_{i=1}}\log (1 + \tilde{\tau}_i\sigma_{\wo i}^2) 
+ \frac{1}{2}\overset{m}{\sum_{i=1}}\frac{\mu_{\wo i}^2\tilde{\tau}_i - 2\mu_{\wo i}\tilde{\nu}_i - \tilde{\nu}^{2}_i\sigma^2_{\wo i}}{(1 + \tilde{\tau}_i\sigma_{\wo i}^2) }, 
\end{eqnarray}

\noindent where the second equation is just expanding the $\log \tilde{Z}_i$ terms, the fourth equation combines the second and eighth terms of the previous equation, the fifth equation combines the fifth and the eighth terms of the previous equation, and the final equation rewrites those in terms of natural parameters. These tedious but straightforward steps enable calculation of $\log Z$ using only the natural parameters of the approximate factors.  Doing so yields a numerically stable implementation.  

Finally, we can go a few steps further to simplify the run time and the numerical stability of calculating $\log Z$.  Since we often want to calculate small probabilities with EPMGP, and do so quickly, these extra steps are important.  Noting the definition of $\Sigma$ (as in Equation \ref{eqn:epmgppost}), we write:

\begin{eqnarray}
\label{eqn:logZstable}
\log Z &= & 
-\frac{1}{2}\log\lvert K \rvert 
+ \frac{1}{2}\log\lvert \Sigma \rvert 
-\frac{1}{2}\m^TK^{-1}\m \nonumber 
+\frac{1}{2}\boldmu^T\Sigma^{-1}\boldmu
\\ && ~~
+ \overset{m}{\sum_{i=1}}\log
\hat{Z}_i + \frac{1}{2}\overset{m}{\sum_{i=1}}\log (1 + \tilde{\tau}_i\sigma_{\wo i}^2) 
+ \frac{1}{2}\overset{m}{\sum_{i=1}}\frac{\mu_{\wo i}^2\tilde{\tau}_i - 2\mu_{\wo i}\tilde{\nu}_i - \tilde{\nu}^{2}_i\sigma^2_{\wo i}}{(1 + \tilde{\tau}_i\sigma_{\wo i}^2) }, \nonumber \\
&= & 
-\frac{1}{2}\log\lvert K \rvert 
- \frac{1}{2}\log\Bigl\lvert     \Bigl(K^{-1} + \overset{m}{\sum_{i=1}} \tilde{\tau}_i\c_i\c_i^T\Bigr)  \Bigr\rvert 
-\frac{1}{2}\m^TK^{-1}\m \nonumber 
+\frac{1}{2}\boldmu^T\Bigl( K^{-1} + \overset{m}{\sum_{i=1}} \tilde{\tau}_i\c_i\c_i^T\Bigr ) \boldmu
\\ && ~~
+ \overset{m}{\sum_{i=1}}\log
\hat{Z}_i + \frac{1}{2}\overset{m}{\sum_{i=1}}\log (1 + \tilde{\tau}_i\sigma_{\wo i}^2) 
+ \frac{1}{2}\overset{m}{\sum_{i=1}}\frac{\mu_{\wo i}^2\tilde{\tau}_i - 2\mu_{\wo i}\tilde{\nu}_i - \tilde{\nu}^{2}_i\sigma^2_{\wo i}}{(1 + \tilde{\tau}_i\sigma_{\wo i}^2) }, \nonumber \\
&= & 
-\frac{1}{2}\log\Bigl\lvert     \Bigl(I + \overset{m}{\sum_{i=1}} \tilde{\tau}_i(L^T\c_i)(L^T\c_i)^T\Bigr)  \Bigr\rvert 
-\frac{1}{2}\m^TK^{-1}\m 
+\frac{1}{2}\boldmu^TK^{-1}\boldmu 
+\frac{1}{2}\overset{m}{\sum_{i=1}} \tilde{\tau}_i(\c_i^T\boldmu)^2 \nonumber
\\ && ~~
+ \overset{m}{\sum_{i=1}}\log
\hat{Z}_i + \frac{1}{2}\overset{m}{\sum_{i=1}}\log (1 + \tilde{\tau}_i\sigma_{\wo i}^2) 
+ \frac{1}{2}\overset{m}{\sum_{i=1}}\frac{\mu_{\wo i}^2\tilde{\tau}_i - 2\mu_{\wo i}\tilde{\nu}_i - \tilde{\nu}^{2}_i\sigma^2_{\wo i}}{(1 + \tilde{\tau}_i\sigma_{\wo i}^2) },
\end{eqnarray}

\noindent where $K = LL^T$ is the Cholesky factorisation of the prior covariance $K$.  We can precompute $L$ at the beginning of the algorithm, and hence the terms $\m^TK^{-1}\m$ and $\boldmu^TK^{-1}\boldmu$ are also easy to solve.  Then, all terms are highly numerically stable, and only that first log determinant has cubic run time.  Importantly, we never have to invert or solve $\Sigma$, and we are able to operate solely in the space of natural parameters for the factor approximations.

To sum up, these steps give us fast and numerically stable implementations for calculating the updates to $q(\x)$ and for calculating the final normalising constant, which is the result of our EPMGP algorithm.   Our simple MATLAB implementation can be downloaded from {\tt <url to come with publication>}.

\newpage

\section{Derivatives of the EP result $\log Z$}
\label{sec:deriv}

The main text has claimed that an advantage of EP methods is that they provide an analytical approximation to the Gaussian (log) probability $\log Z$ such that derivatives can be taken with respect to parameters of the Gaussian $p_0(\x) = \mathcal{N}(\x; \m, K)$.  We produce those derivatives here as reference.  Throughout applied uses of EP, derivatives with respect to $\log Z$ are often interesting, as $\log Z$ in a standard approximate inference model corresponds to the marginal likelihood of the observed data. One often wants to optimise (and thus take derivatives of) this quantity with respect to parameters of the prior $p_0(\x)$.  Two useful references for discussion and application of these derivatives are \cite{seeger08epexpfam, rasmussenBook}.  Most importantly, we leverage the result from \cite{seeger08epexpfam} (Section 2.1), where it is shown that the converged messages - the approximate factor parameters $\{\tilde{Z}_i,\tilde{\mu}_i,\tilde{\sigma}_i^2\}$ - have 0 derivative with respect to parameters of the prior $p_0(\x)$.  This consideration is sometimes described as a distinction between \emph{explicit} dependencies (where $\log Z$ actually has the prior terms $\{\m,K\}$) and \emph{implicit} dependencies where a message parameter such as $\tilde{Z}_i$ might depend implicitly on $\{\m,K\}$ (as in \cite{rasmussenBook}).  The fact that EP converges to a fixed point of its objective function, and thus makes the latter implicit derivatives equal 0, simplifies our task considerably. 

The simplest place to begin is with the original form of the normaliser in Equation \ref{eqn:logZ}.  We note that the second line of that equation is entirely due to the approximate factors, so by \cite{seeger08epexpfam}, we need only consider the first and third lines, yielding:

\begin{equation}
\label{eqn:dlogZ}
\log Z  ~~~ = ~~~ 
\frac{1}{2}\boldmu^T\Sigma^{-1}\boldmu
+ \frac{1}{2}\log\lvert \Sigma \rvert
-\frac{1}{2}\m^TK^{-1}\m 
-\frac{1}{2}\log\lvert K \rvert  + \mathrm{constants}.
\end{equation}

Typically, these terms can be simplified into the familiar form of a normaliser of a product of Gaussians ({\it e.g.} A.7 in \cite{rasmussenBook}).   However, because the factor contributions need not be full rank, that standard treatment requires a bit more detail.  First, we revisit the definition of $\{\boldmu, \Sigma\}$:

\begin{equation}
\boldmu  =
\Sigma\Bigl(K^{-1}\m + \r \Bigr),
~~~~~~\Sigma = \Bigl(K^{-1} + RR^T \Bigr)^{-1},
\end{equation}

\noindent where we have hidden several of the approximate factor parameters with:

\begin{equation}
\label{eqn:bterms}
\r  = \overset{m}{\sum_{i=1}}\frac{\tilde{\mu}_i}{\tilde{\sigma}^2_i}\c_i,
~~~~~~
RR^T = \overset{m}{\sum_{i=1}} \frac{1}{\tilde{\sigma}^2_i}\c_i\c_i^T.
\end{equation}

\noindent $RR^T$ is a singular value decomposition of the matrix above (in general $R$ will not be square).  This simplification allows us to focus only on the parts of the equation that are dependent on $\{\m,K\}$.  Thus we can rewrite $\log Z$ as:

\begin{eqnarray}
\label{eqn:dlogZ2}
\log Z  ~~~&=&~~~ 
-\frac{1}{2}\log\lvert K \rvert 
+ \frac{1}{2}\log\lvert (K^{-1} + RR^T)^{-1} \rvert + \mathrm{constants}
\\ && ~~~~~
+ \frac{1}{2}(K^{-1}\m + \r)^T(K^{-1} + RR^T)^{-1}(K^{-1}\m + \r)
-\frac{1}{2}\m^TK^{-1}\m
\nonumber
\\
~~~& = & ~~~
-\frac{1}{2}\log\lvert (I + R^TKR) \rvert + \mathrm{constants}
\\ && ~~~~~
-\frac{1}{2} (\m -K\r)^T\Bigl(R(I + R^TKR)^{-1}R^T\Bigr)(\m - K\r) + \r^T(\m + \frac{1}{2}K\r)
\nonumber
\end{eqnarray}

\noindent where the second line comes from combining the two determinants and using Sylvester's determinant rule, and the third line comes from the typical algebra of using the matrix inversion lemma to rearrange these terms.  Some readers may prefer the second or the third line;  we use the third as it prevents any explicit inversions of $K$ and allows significant computational savings if the number of columns of $R$ is smaller than that of $K$.

With that setup, we consider two possible and common derivative quantities of interest.  First and most simply, we may want the gradient with respect to the Gaussian mean $\m$:

\begin{equation}
\nabla_{\m} \log Z = - \Bigl(R(I + R^TKR)^{-1}R^T\Bigr)(\m - K\r) + \r
\end{equation}

\noindent This form has a closed form solution and is instructive for two reasons.  First, we note that if $RR^T$ is not full rank, then $\m$ is underdetermined.  This is intuitively reasonable, as a subrank $RR^T$ means that the polyhedron is open in some dimension ({\it e.g.} a slab), and thus there should be a linear subspace of optimal positions for the Gaussian.  Second, this solution is interesting in its connections to centering algorithms as previously mentioned (see for example Appendix \ref{sec:minpoly}).  This optimised mean provides a center for a $K$ shaped Gaussian within the polyhedron $\regionA$.  Understanding the geometric implications of this centering approach is an interesting question, but well beyond the scope of this work. 

Second,  we can consider parameterised $K$.  We presume there is some parameter $\theta$ (such as a kernel hyperparameter) on the covariance $K = K(\theta)$.  Then we see: 

\begin{eqnarray}
\label{eqn:dlogZ3}
\frac{\partial \log Z}{\partial \theta}
~~& = & ~~~
\frac{\partial}{\partial \theta} \Biggl[ -\frac{1}{2}\log\lvert (I + R^TKR) \rvert    \nonumber
\\ && ~~~~~~~~~~
-\frac{1}{2} (\m -K\r)^T\Bigl( R(I + R^TKR)^{-1}R^T\Bigr)(\m - K\r) + \r^T(\m + \frac{1}{2}K\r) \Biggr]
\nonumber
\\
&& \nonumber
\\
~~& = & ~~~
-\frac{1}{2}\text{trace}\left( R(I + R^TKR)^{-1}R^T\frac{\partial K}{\partial \theta} \right)   \nonumber
\\ && ~~~
+ \r^T \frac{\partial K}{\partial \theta} \Bigl(R(I + R^TKR)^{-1}R^T\Bigr)(\m - K\r)  \nonumber
\\ && ~~~
+ \frac{1}{2} (\m -K\r)^T\Bigl(R(I + R^TKR)^{-1}R^T\frac{\partial K}{\partial \theta}R(I + R^TKR)^{-1}R^T\Bigr)(\m - K\r)  \nonumber
\\ && ~~~
+ \frac{1}{2}\r^T\frac{\partial K}{\partial \theta}\r.
\end{eqnarray}

\noindent Within this form, the matrix $R(I + R^TKR)^{-1}R^T$ reverts to the familiar inverse sum of covariances when $R$ is full rank.  Thus, though this form may look different, it is simply a subrank version of the familiar product of Gaussians marginalisation constant as often used in marginal likelihood derivatives for EP.

Finally, we note that we have only focused on derivatives with respect to parameters of the Gaussian $p_0(\x)$.  One might also want to take parameters with respect to the region $\regionA$.  The convenience of ignoring implicit dependencies would no longer hold in such a case, so taking these derivatives would involve taking more complicated derivatives of the approximate factors.  As this is an aside to the current report, we leave that investigation for future work.

\newpage

\section{Minimalising polyhedral regions $\regionA$}
\label{sec:minpoly}

Our box-function definition of polyhedra, or similarly the more conventional halfspace definition, does not in general uniquely specify a polyhedral region $\regionA$.   For any polyhedron defined by $\regionA = \prod_i t_i(\x)$ (our conventional definition throughout the paper), we can always add another constraint $t_j(\x)$ such that $\regionA = \prod_i t_i(\x) = t_j(\x) \prod_i t_i(\x)$.   This redundant constraint could be colinear with an existing $t_i(\x)$ factor that includes at least the $[l_i,u_i]$ interval, but it can also be any constraint such that the entire projection of $\regionA$ onto $\c_j$, the rank one dimension of $t_j(\x)$, is contained inside $[l_j,u_j]$.  Such factors are typically called inactive constraints.  

Importantly, these inactive constraints do not change the polyhedron and therefore do not change the probability $F(\regionA)$ of any Gaussian across this region.  On the other hand, inactive constraints do change the representation and thus the EP approximation.  Given our discussion of the extra mass and redundancy issues (Sections \ref{sec:results} and \ref{sec:discussion}), removing these inactive constraints can potentially improve the accuracy of our EP algorithms.  To do so, we define a minimal representation of any polyhedron $\regionA$ to be the polyhedron where all constraints are active and unique.  This can be achieved by pulling in all inactive constraints until they support the polyhedron $\regionA$ and are thus active, and then we can prune all repeated/redundant constraints.  These steps are solvable in polynomial time as a series of linear programs, and they bear important connection to problems considered in centering (analytical centering, Chebyshev centering, etc.) for cutting-plane methods and other techniques in convex optimisation ({\it e.g.}, see \cite{boydBook}).    The first problem one might consider is to find a point on the interior of the polyhedron, which can be solved by introducing a slack variable $t$ into the linear program:  

\begin{equation}
\label{eqn:lp0}
\begin{split}
\underset{\{\x,t\}}{\mathrm{minimise}} & ~~~~t   \\
\text{subject\thinspace to} & ~~~-\c_i^T\x + l_i  <  t ~~\forall~ \text{$i$ = 1{\ldots}$m$}\\
& ~~~~~~\c_i^T\x - u_i  <  t ~~\forall~ \text{$i$ = 1{\ldots}$m$}.
\end{split}
\end{equation}

This program will either return an interior point $\x$ and a negative slack $t$ or will return a positive $t$, which is a certificate that the polyhedron is in fact empty.  Standard LP solvers can be used to find this result.  Next, we can consider each factor $t_j(\x)$ to test if it is inactive.  First to consider the lower bound, for all $j$ we solve: 

\begin{equation}
\label{eqn:lp1}
\begin{split}
\underset{\x}{\mathrm{minimise}} & ~~~~\c_j^T\x   \\
\text{subject\thinspace to} & ~~~-\c_i^T\x + l_i  <  t ~~\forall~ \text{$i$ = 1{\ldots}$m$}\\
& ~~~~~~\c_i^T\x - u_i  <  t ~~\forall~ \text{$i$ = 1{\ldots}$m$}.
\end{split}
\end{equation}

\noindent Having solved for the lower bounds, we then solve the upper bound problems for all $j$: 

\begin{equation}
\label{eqn:lp2}
\begin{split}
\underset{\x}{\mathrm{maximise}} & ~~~~\c_j^T\x   \\
\text{subject\thinspace to} & ~~~-\c_i^T\x + l_i  <  t ~~\forall~ \text{$i$ = 1{\ldots}$m$}\\
& ~~~~~~\c_i^T\x - u_i  <  t ~~\forall~ \text{$i$ = 1{\ldots}$m$}.
\end{split}
\end{equation}

With the answers for these $2m$ LPs, we can now make all constraints active.  If the solution to Equation \ref{eqn:lp1} is larger than $l_j$, then we know that the lower bound $l_j$ is inactive.  In words, this means that the other constraints have cut off $\regionA$ inside of $l_j$ in direction $\c_j$.  Equivalently, if the solution to Equation \ref{eqn:lp2} is smaller than the upper bound $u_j$, we know $u_j$ is inactive.  If both lower and upper bounds are found to be inactive, the factor $t_j(\x)$ is entirely redundant and can be pruned from the representation without changing $\regionA$.  Since we know redundancy can lead to EP underestimates, pruning can be useful (see for example Panel A of Figure \ref{fig:resultspathology}).  If only one of the lower or upper bounds is found to be inactive by Equation \ref{eqn:lp1} or \ref{eqn:lp2}, then this bound can be tightened to {\it activate} it, at which point it will support the polyhedron $\regionA$.   The value it should be tightened to is the result of the corresponding LP that found that bound to be inactive.  This tightening operation would prevent errors such as those seen in Panel B of Figure \ref{fig:resultspathology}.  One small nuisance is simple repeated active factors, as each will technically appear active.  However, this case can be dealt with by checking for such repetitions in preprocessing, and thus is not a difficulty in practice.  

With these steps done, we have a minimal representation of the polyhedron $\regionA$.  This set of operations is convenient because using it as a preprocessing step will ensure that each polyhedron is uniquely represented, and thus results from EP and other such approaches (as with centering algorithms) are consistent.  However,  in our results and our presentation we do not minimalise polyhedra to avoid adding additional complexity.  Furthermore, as in Figure \ref{fig:resultspathology}, doing so can mask other features of EP that we are interested in exploring.  Anecdotally, we report that while minimalising polyhedra \emph{will} change EP's answer, the effect is not particularly large in these randomly sampled polyhedral cases.

\newpage

\section{Minimising KL-divergence does moment matching}
\label{sec:KL}

In this section, we review the equivalence between moment matching and minimising the unnormalised KL-divergence in an exponential family. This fact is often mentioned without proof in the literature (as for example in~\cite{minkaMSFTTR2005}) -- we provide here the derivation for completeness. In particular, this implies that minimising the KL-divergence corresponds to matching the zeroth, first, and second moments of $q(\x)$ to $p(\x)$, when the approximating family is Gaussian.  As approximately minimising the KL-divergence is the goal of EP, this result will show that EPMGP is an appropriate choice to solve the Gaussian probability calculation (zeroth moment). We provide the general derivation here.

First, as defined in the main text, $p(\x)$ and $q(\x)$ do not normalise to 1. Thus, we use the general definition of the KL-divergence for non-negative unnormalised distributions $p(\x)$ and $q(\x)$:
\begin{equation}
\label{eqn:KLfg}
D_{KL}\bigl(p(\x)\parallel q(\x)\bigr) = \int p(\x) \log \frac{p(\x)}{q(\x)} d\x
~~+~~\int q(\x)d\x ~~-~~ \int p(\x)d\x,
\end{equation}
as in \cite[]{zhu95infogeom, minkaMSFTTR2005}.  Note that the typically-seen normalised KL-divergence ({\it e.g.}, \cite{CoverandThomas}) is recovered when both $p(\x)$ and $q(\x)$ normalise to 1.

We are interested in the distribution $q(\x)$ that minimises $D_{KL}\bigl(p(\x) \parallel q(\x)\bigr)$ when $q(\x)$ is restricted to be in an (unnormalised) exponential family, and where $p(\x)$ could be arbitrary. The exponential families are flexible parametric families of distributions which contain many familiar families such as the Gaussian, Dirichlet, Poisson, etc.~\cite[]{bishopBook,seeger03phd}. Reusing the notation from~\cite{seeger03phd}, an exponential family is defined by specifying the sufficient statistics vector function $\phi(\x)$\footnote{Strictly speaking, the base measure $\mu$ for the density also needs to be specified (e.g. the counting measure for a Poisson distribution). For simplicity of presentation, we will show our derivations with the Lebesgue measure (appropriate for the Gaussian distribution e.g.), though the result holds for any exponential family in the natural parameterisation.}, and so the unnormalised $q(\x)$ takes the form:
\begin{equation}
\label{eqn:expfam2}
q(\x) = Z_q\exp\Bigl\{\boldtheta^T\phi(\x) -
\Phi(\boldtheta)\Bigr\},
\end{equation}
where $\Phi(\boldtheta) =
\log\int\exp\Bigl\{\boldtheta^T\phi(\x)\Bigr\}d\x$ is the log-partition function (the usual normaliser) and $Z_q$ is an extra parameter accounting for the unnormalised $q(\x)$. Substituting this form of $q(\x)$ in~\eqref{eqn:KLfg}, we obtain:
\begin{align}
\label{eqn:KLpq}
D_{KL}\bigl(p(\x) \parallel q(\x)\bigr)
& = \int p(\x) \left(\Phi(\boldtheta) - \boldtheta^T\phi(\x) - \log{Z_q} \right)d\x +Z_q-\int p(\x)d\x - H[p] \\
& = Z_p \Phi(\boldtheta) - \boldtheta^T \int \phi(\x) p(\x) d\x  - Z_p \log{Z_q} + (Z_q-Z_p) - H[p],
\end{align}
where $H[p] = -\int p(\x) \log{p(\x)} d\x$ is the entropy of $p$ (a constant with respect to $q$), and we defined the normalisation constant of $p$ as $Z_p = \int p(\x) d\x$. To minimise this KL, we
first take the derivative with respect to the normaliser of $q(\x)$,
namely $Z_q$:
\begin{eqnarray}
\label{eqn:dZ}
\frac{d}{dZ_q}D_{KL}\bigl(p(\x) \parallel q(\x)\bigr)
& = & -~\frac{Z_p}{Z_q} + 1 = 0 \\
\label{eqn:Zstar}
& \implies & Z_q^* = Z_p = \int p(\x)d\x.
\end{eqnarray}
The second derivative is $1/Z_q^2 > 0$, so this is the unique global minimum.

From this, we see that minimising the KL-divergence with respect to the normalising constant $Z_q$ matches the zeroth moments, as we expected.  Indeed, it is this equation that motivates the entire EPMGP approach. We obtain the optimal natural parameter $\boldtheta^*$ of $q(\x)$ by setting its derivative of the KL to zero:
\begin{eqnarray}
\label{eqn:dtheta}
\frac{d}{d\boldtheta}D_{KL}\bigl(p(\x) \parallel q(\x)\bigr)
& = & Z_p \nabla_{\boldtheta}\Phi(\boldtheta)- \int p(\x)\phi(\x)d\x = \mathbf{0} \\
& \implies & \nabla_{\boldtheta}\Phi(\boldtheta) = \int \frac{p(\x)}{Z_p}\phi(\x)d\x \label{eqn:p-moment}.
\end{eqnarray}
It is a standard fact of the exponential family that the gradient of the log-partition function is equal to the expected sufficient statistics; to see this, from the definition of $\Phi(\cdot)$ (after Equation~\ref{eqn:expfam2} above), we have:
\begin{eqnarray}
\label{eqn:dphi}
\nabla_{\boldtheta}\Phi(\boldtheta)
& = & \frac{\int\phi(\x)\exp\Bigl\{\boldtheta^T\phi(\x)\Bigr\}d\x}
{\int\exp\Bigl\{\boldtheta^T\phi(\x)\Bigr\}d\x} \\
& = & \int\phi(\x)\exp\Bigl\{\boldtheta^T\phi(\x) -
\Phi(\boldtheta)\Bigr\}d\x \\ \label{eqn:Eqphi}
& = & \int \frac{q(\x)}{Z_q}\phi(\x)d\x.
\end{eqnarray}
Combining~\eqref{eqn:Eqphi} with~\eqref{eqn:p-moment}, we get that the expected sufficient statistics of $q$ need to match those of $p$, namely, the higher-order moment matching condition, as we wanted to show. The second derivative with respect to $\boldtheta$ of the KL-divergence is the Hessian of the log-partition function, which is strictly positive definite if the exponential family is \emph{minimal}~\cite[]{wainwright08variational} and in this case gives a unique $\boldtheta^*$; otherwise, the moment matching condition still gives a global minimum, but there can be multiple parameters with this property.  For the Gaussian family, the sufficient statistics are a vector of all elements $x_i$ and all pairs of elements $x_i x_j$. This means that the moment matching condition is to match the zeroth, first and second order statistics.

To summarise, these final equations~\eqref{eqn:Zstar}, \eqref{eqn:Eqphi} and ~\eqref{eqn:p-moment} tell us that, to uniquely minimise the global KL-divergence between the truncated distribution $p(\x)$ (or any target distribution) and the Gaussian $q(\x)$, we do two things: first, we set $Z$ to be the total mass (zeroth moment) of $p(\x)$; and second, we set $\boldtheta$, the natural parameters of $q(\x)$ such that the mean (first moment) and covariance (second moment) of $q(\x)$ equal exactly the first and second moments of $p(\x)$. 

As it pertains to calculating Gaussian probabilities, minimising the KL-divergence calculates the zeroth moment of $p(\x)$, which is exactly $F(\regionA)$, the probability of interest.  As such, EPMGP is a sensible choice for multivariate Gaussian probability calculations.

\newpage

\bibliography{epmgp}

\end{document}